\documentclass[journal]{IEEEtran}
\usepackage[hidelinks]{hyperref}
\usepackage{graphicx,amsmath,amssymb,soul,tabu,epstopdf,mathtools,multirow,booktabs}
\usepackage[symbol]{footmisc}
\usepackage{subcaption}	
\usepackage[framemethod=TikZ]{mdframed}
\usepackage{algorithm,algpseudocode}
\usepackage{array}
\usepackage{amsthm}
\usepackage{paralist}

\newtheorem{definition}{Definition}    %% this does it
\newcolumntype{P}[1]{>{\centering\arraybackslash}p{#1}}

\newcommand{\etal}{\MakeLowercase{\textit{et al.}}}
\newcommand{\mb}[1]{{\boldsymbol{#1}}}
\newcommand{\ty}[1]{{\scriptscriptstyle{\mathcal{#1}}}}
\newcommand{\trsp}{\mathsf{T}}
\newcommand{\psin}{{\dagger}}
\newcommand{\HLeft}{\overset{+}{\mb{H}}}

\newcommand{\HRightVec}{\bar{\mb{H}}^{*}}

\newcolumntype{C}[1]{>{\centering\let\newline\\\arraybackslash\hspace{0pt}}m{#1}}	% to center text in fixed-width columns
\newcommand*\rot{\rotatebox{90}}	% to rotate text in a box

\begin{document}
\title{Learning Task Priorities from Demonstrations}

\author{Jo\~ao~Silv\'erio,~ Sylvain~Calinon,~ Leonel~Rozo,~ Darwin~G.~Caldwell%\IEEEmembership{Member,~IEEE}
	\thanks{J. Silv\'erio, L. Rozo and D. G. Caldwell are with the Department of Advanced Robotics, Istituto Italiano di Tecnologia, 16163 Genova, Italy (e-mail: joao.silverio@iit.it; leonel.rozo@iit.it; darwin.caldwell@iit.it).}
	\thanks{S. Calinon is with the Idiap Research Institute, CH-1920 Martigny, Switzerland, and with the Department of Advanced Robotics, Istituto Italiano di Tecnologia, 16163 Genova, Italy (e-mail: sylvain.calinon@idiap.ch).}
	\thanks{This work was supported by the MEMMO project (European Union's Horizon 2020 Programme, Grant 780684).}
	\thanks{We would like to thank Luca Muratore and Phil Hudson for assisting with the COMAN experiments and Dr Martijn Zeestraten, Arturo Laurenzi and Giuseppe Rigano for the help with the Centauro simulator. We would also like to thank Dr Yanlong Huang for his feedback on previous versions of the paper.}
}

\markboth{IEEE Transactions on Robotics}%
{IEEE Transactions on Robotics}
\maketitle

\begin{abstract}
Bimanual operations in humanoids offer the possibility to carry out more than one manipulation task at the same time, which in turn introduces the problem of task prioritization. We address this problem from a learning from demonstration perspective, by extending the Task-Parameterized Gaussian Mixture Model (TP-GMM) to Jacobian and null space structures. The proposed approach is tested on bimanual skills but can be applied in any scenario where the prioritization between potentially conflicting tasks needs to be learned. We evaluate the proposed framework in: two different tasks with humanoids requiring the learning of priorities and a loco-manipulation scenario, showing that the approach can be exploited to learn the prioritization of multiple tasks in parallel. 
\end{abstract}


\graphicspath{{./figures/}}
\section{Introduction}

\IEEEPARstart{T}{he} human-robot transfer of bimanual skills is a growing topic of research in robotics. As the number of available dual-arm platforms and humanoid robots increases, existing learning and control algorithms must be reformulated to accommodate the constraints imposed by these morphologies and to take full advantage of the repertoire of tasks that such robots can perform \cite{Smith2012}. %Among the available learning techniques,
Learning from Demonstration (LfD) \cite{Billard2008} is a particularly promising direction to achieve a seamless transfer of bimanual abilities, but it has so far mostly addressed the  %extracting invariant patterns in
learning of uni-manual and single-task skills.
This article aims at extending the LfD paradigm to the learning of elaborated features that arise during bimanual manipulation in humanoids, particularly task prioritization.%hierarchical task prioritization, including the simultaneous handling of configuration and operational space constraints. 

\begin{figure}
	\centering
	\includegraphics[width=0.99\columnwidth]{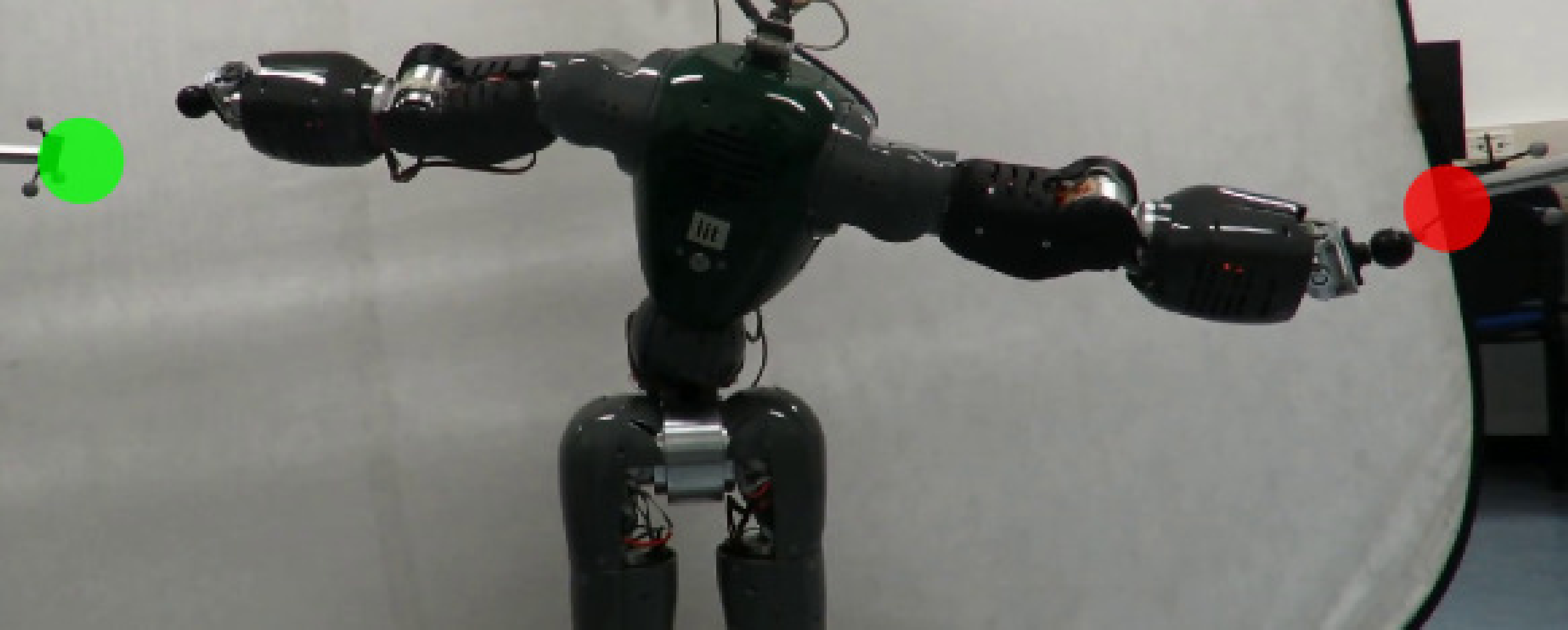}\\
	\vspace{0.1cm}
	\includegraphics[width=0.99\columnwidth]{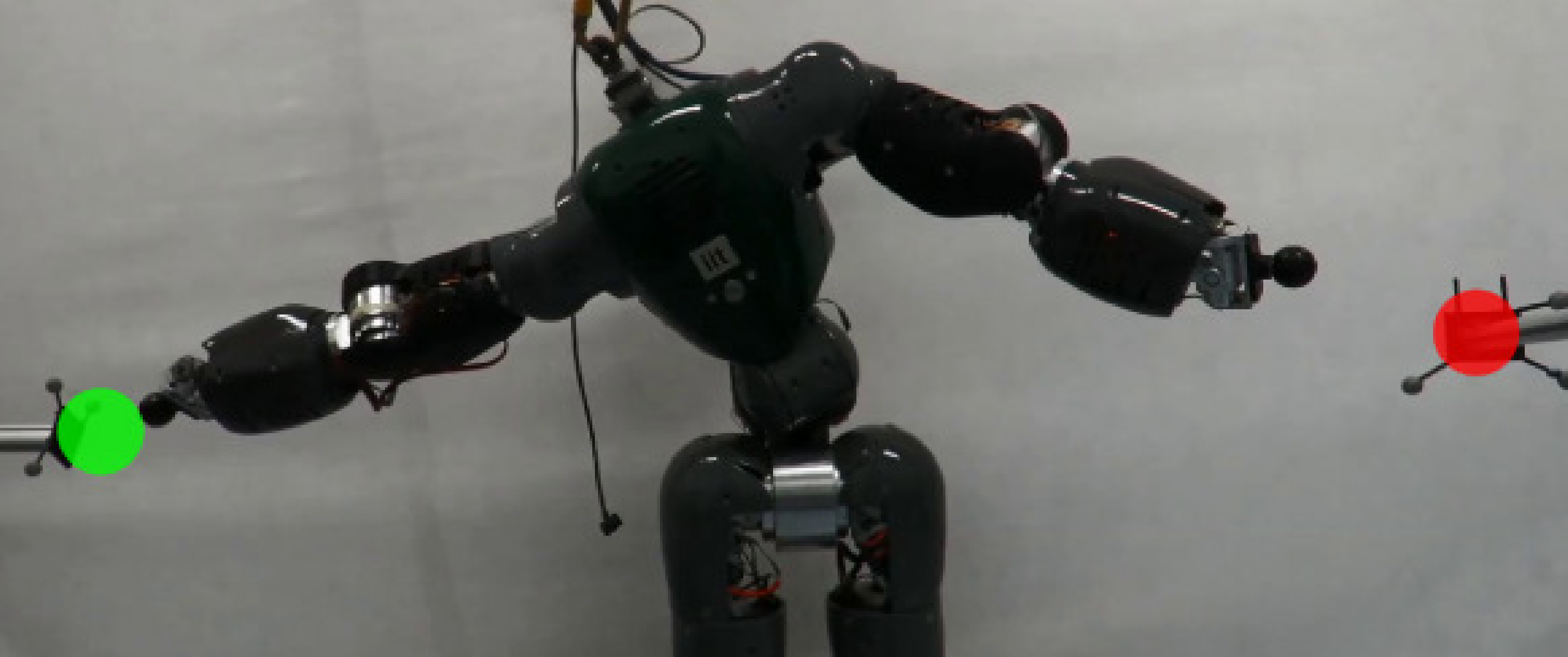}
	\caption{The COMAN robot prioritizes the tracking of the left (resp.\ right) target, while doing its best to track the right (resp.\ left) one. This priority behavior was learned from demonstrations, using the approach proposed in Section \ref{sec:priorityConstraints}. \textbf{Top:} Reproduction with the model trained on left arm priority demonstrations. \textbf{Bottom:} Reproduction with the model trained on right arm priority demonstrations.}
	\label{fig:COMANPriority}
	\vspace{-0.1cm}
\end{figure}
As humans, we employ rich bimanual coordination behaviors on a daily basis (e.g., tying knots, moving heavy or bulky objects, sweeping). %, as well as complex decision-making such as which arm should be favored when the two must handle incompatible tasks.
For this reason, most research on bimanual skill learning exploits operational space formulations (e.g. \cite{Gams14TRO,Umlauft14ICRA,Lioutikov2014,Likar15,Ureche14,Calinon12Humanoids,Silverio15}), that focus on task space \textit{constraints}, e.g. demonstrated coordination between end-effectors and object-related movements, that need to be reproduced precisely in order to successfully complete a task. However, constraints also arise in configuration space, for example as preferred body/arm postures or movements for which joint trajectories are more important than those of end-effectors. In such scenarios, operational space formulations alone are insufficient for correct task execution.
Similarly, humanoids are often required to perform dexterous dual-arm skills that demand handling multiple potentially conflicting tasks in parallel. These conflicts can occur at various levels, such as when determining how to use the torso joints if both arms are needed (Fig.\ \ref{fig:COMANPriority}), or how to switch between poses while keeping balance. 
Endowing robots with the ability to learn how to handle priorities is an important research problem. It relates to the challenge of organizing movement primitives not only in series but also in parallel, which is often overlooked.

In this article, we offer a new perspective on task-parameterized movement models by including Jacobian matrices and null space structures in their formulation. Our approach takes Task-Parameterized Gaussian Mixture Models (TP-GMM) \cite{Calinon12Humanoids,Calinon16JIST} (reviewed in Section \ref{sec:LearningTPMovements}) as an example, which was originally used to probabilistically encode Cartesian end-effector motions from the perspective of different coordinate systems. %in the workspace of the robot
%representing object poses. % or landmarks.
We propose here a more general formulation, exploiting the affine operations structure of TP-GMM to address the aforementioned LfD problems. %In particular, we consider operators based on Jacobian and null space projection matrices that allow for solving the aforementioned LfD problems.
More specifically, the contribution of this paper is a novel formulation of TP-GMM to:
\begin{compactenum}
	\item \textbf{Simultaneously learn constraints in operational and configuration spaces (Section \ref{sec:combiningConstraints}).} This article improves on previous work \cite{Calinon09AR} in two directions: i) it formalizes the handling of operational and configuration space constraints in the context of TP-GMM and ii) it introduces unit quaternion-based projection operators, that permit the learning of orientation constraints.
	\item \textbf{Learn task prioritization hierarchies (Section \ref{sec:priorityConstraints}).} Our formulation permits the identification of demonstrated priority behaviors, given an initial set of \textit{candidate} task hierarchies. In addition, it allows the robot to reproduce the learned priorities in new situations. To the best of our knowledge, this is the first approach that permits learning full hierarchy structures from only few demonstrations.
\end{compactenum}

Some of the aforementioned points were briefly introduced in \cite{Calinon16JIST}. Here we provide an extensive derivation, analysis and validation in \iffalse two \fi three experimental scenarios, with real and simulated robotic platforms. First, we use the COMpliant huMANoid (COMAN) \cite{Tsagarakis13} robot to show that TP-GMM can be used to handle operational and configuration space constraints simultaneously. For this, we choose the skill of bimanually shaking a bottle (Section \ref{subsec:shakingTask}). We then use a bimanual reaching task to teach priorities from demonstrations (Section \ref{sec:expCOMAN}). The task consists of tracking two conflicting targets on the left and right sides of COMAN with the corresponding arm. Finally, we consider a loco-manipulation scenario with the Centauro robot \cite{Baccelliere2017} in simulation (Section \ref{sec:expCentauro}). In this experiment, the prioritization of floating base position, end-effector positions and end-effector orientations needs to be learned, showing that the approach can be exploited in generic task prioritization applications, in particular with hierarchies of more than two tasks.

%%%%%%%%%%%%%%%%%%%%%%%%%%%%%%%%%%%%%%%%
% II. Related Work
%%%%%%%%%%%%%%%%%%%%%%%%%%%%%%%%%%%%%%%%

\section{Related Work}
\label{sec:relatedWork}

\subsection{Learning bimanual skills}
The most popular approaches for learning bimanual manipulation from demonstrations are based on Dynamic Movement Primitives (DMPs) \cite{Ijspeert13}. Examples range from the use of virtual springs between end-effectors \cite{Gams14TRO} to the coupling of DMPs using artificial potential fields \cite{Umlauft14ICRA}. Lioutikov \etal\ \cite{Lioutikov2014} propose to combine sequences of DMPs that encode partial demonstrations of complete individual arm movements. Similarly to the spirit of DMPs, Likar \etal\ \cite{Likar15} introduce an approach based on Iterative Learning Control for force adaptation in bimanual tasks. In a more probabilistic fashion, Ureche and Billard \cite{Ureche14} focus on the extraction of arm dominance and role from demonstrations, as well as on the correlations between task variables such as poses and forces. Our previous work \cite{Calinon12Humanoids, Silverio15} follows a task-parameterized approach to learning bimanual skills, where relative and absolute end-effector movements are encoded with respect to a pre-defined set of coordinate systems, whose importance is learned probabilistically from demonstrations.

The foregoing collection of work addresses bimanual skill transfer from an operational space perspective. Consequently, rich features % with several degrees of freedom (DOFs),
such as joint space movements and task prioritization,  commonplace in highly redundant manipulators, cannot be adequately learned.

\subsection{Simultaneous learning of operational and configuration space constraints}

The problem of knowing which space---between configuration and operational spaces---is the most relevant for a given task, has frequently been treated as a hand-tuning of scalar weights assigned to sub-tasks in each space \cite{Moro13}. Exceptions include frameworks based on reinforcement learning (RL) \cite{Dehio15,Modugno16} and LfD \cite{Calinon09AR}, where the importance of each space is learned. Approaches like \cite{Dehio15,Modugno16} employ stochastic optimization, given a set of reward functions related to high level goals, to find optimal weights. In \cite{Calinon09AR}, the authors treat the problem as a weighted least squares problem where the weights associated with each space, encoded as full precision matrices, reflect the variability and correlations in the demonstrations. The two types of approach are similar in concept, with \cite{Calinon09AR} and \cite{Dehio15,Modugno16} exploiting velocity and torque controllers, respectively. % to map operational space movements onto configuration space
Despite the similarities, the RL-based approaches require that considerable prior knowledge is accounted for in the reward functions, making those methods less straightforward in many applications. In Section \ref{sec:combiningConstraints} we propose an approach %for simultaneously learning operational and configuration space constraints which is
related to \cite{Calinon09AR}. It improves on that work by considering an arbitrary number of tasks in operational space, as well as orientation constraints, which were previously overlooked. 

Managing constraints from different spaces using full weight matrices, as we do in Section \ref{sec:combiningConstraints}, can be seen as a form of task prioritization---that we refer to as a \textit{fusion of tasks}. It consists of normally distributed control references that are fused as a product of Gaussians. This approach improves the traditional \textit{soft weighting of tasks} (see Section \ref{sec:learningTaskPrioRW}), where scalar weights are used to combine tasks with smooth transitions. The use of full matrices instead of scalar weights provides higher flexibility in the prioritization structure, while maintaining smoothness properties.
Such approaches typically excel in tasks where the different constraints are activated sequentially, as we shall see in Section \ref{sec:combiningConstraints}. Despite the advantages, they tend not to perform well when several constraints are activated at the same time, with similar weights, for which more elaborated prioritization structures are required. In the following section, we review different state-of-the-art techniques for learning task prioritization.

\subsection{Learning task prioritization}
\label{sec:learningTaskPrioRW}

Common approaches to control task prioritization can be categorized in two main directions, by either exploiting \textit{strict hierarchy} structures, applied to multi-level hierarchies \cite{Khatib87,Sentis05,Ott15}, or by employing a \textit{soft weighting} of tasks \cite{Moro13,Park01}. The two techniques have pros and cons. Setting an explicit null space structure guarantees strict priorities at the expense of constraining the tasks, which quickly limits the number of tasks that can simultaneously be handled with the number of degrees of freedom available for controlling the robot. This approach is also prone to discontinuities in the control problem when switching from one hierarchy structure to another. A soft weighting scheme can handle different levels of task importance and gradual changes from one task to another, but it does not provide strict guarantees on the fulfillment of each separated task. 

In an alternative perspective, a collection of work addresses the problem using optimization \cite{Lasa09,Salini11,Morris13} or by leveraging novel representations of prioritization \cite{Dietrich12,Liu16,Dehio2018}. 
In this article, we tackle the challenge from a robot learning perspective. Learning how to handle the priorities of multiple concurrent tasks is a challenging problem in robotics and, despite the recent advancements in this direction, several issues remain open.

\subsubsection{Learning approaches based on strict hierarchies}
Learning priorities based on strict task hierarchies typically assumes that low priority tasks are projected on the null space of high priority ones. In this context, Wrede \etal\ \cite{Wrede13} propose a two-step approach to kinesthetically teach tasks to redundant manipulators. First, in a preliminary phase, the desired null space behavior is demonstrated to the gravity-compensated manipulator. The relation between end-effector positions and desired configurations is encoded in a neural network which is %employed in a subsequent phase to have the robot at demonstrated configurations while the user guides the end-effector to demonstrate the task.
exploited during the demonstrations of the main task to control the robot null space.
Saveriano \etal\ \cite{Saveriano15} propose an approach based on Task Transition Control (TTC) \cite{An15} to refine end-effector and null space policies. They take advantage of the smooth transitions between task priorities allowed by TTC to switch between task execution and teaching, yielding refinement of policies for both the main and the null space tasks in runtime. In \cite{Towell10}, Towell \etal\ aim for extracting underlying null space policies from demonstrations, assuming a strict hierarchy of priorities, but the proposed approach does not allow for the extraction of demonstrated hierarchies. Lin \etal\ \cite{Lin15} propose an approach to learn the kinematic constraints present in movement observations, as explicitly represented by the null space projection matrix of kinematically constrained systems. A common assumption in both \cite{Towell10} and \cite{Lin15} is that one has access to the control variables during the demonstrations. This assumption is shared by our approach (Section \ref{sec:priorityConstraints}), in which we assume to know the control set-point associated with each sub-task.

%These directions of work consider hierarchies of only two tasks -- which significantly limits the potential of their application in humanoid robots -- and tend to require a high number of demonstrations. 

Hak \etal\ \cite{Hak12b} present an iterative algorithm for identifying a \emph{stack of tasks}. From observed joint trajectories, and a pre-defined set of possible tasks that can be executed in parallel, the approach relies on the expected operational space behavior of each task to identify the active ones, and on the task-function formalism to gradually remove tasks \iffalse from the observed movement through null space projections until all tasks have been selected \fi until all have been identified. This approach shares similarities with ours in the assumption that demonstrated movements are generated from a strict hierarchy structure, and in the fact that it also analyses the operational space to disambiguate between possibly active tasks. The main difference is that our task space analysis is probabilistic, while Hak \etal\ rely on a curve fitting score. Moreover, their framework assumes prior knowledge about the possible tasks, while we assume prior knowledge about the possible task hierarchies, including the option to provide an exhaustive list of all possible ones. Finally, our probabilistic formulation permits the learning of input-dependent hierarchies, i.e. the robot can be taught priority behaviors that may vary during reproduction.

\begin{table}
	\centering
	\begin{tabular}{P{2.52cm} P{1.24cm}  P{1.72cm} P{0.3cm}  P{1.05cm}}
		&  Learn hierarchies &  From demonstrations &  2+ tasks & Input dependent\\
		\toprule[1.0pt]
		Dehio \etal\ \cite{Dehio15} & -- & -- & \checkmark & -- \\
		Modugno \etal\ \cite{Modugno16} & -- & -- & \checkmark & \checkmark\\
		Wrede \etal\ \cite{Wrede13} & -- & \checkmark &	-- & --\\
		Hak \etal\ \cite{Hak12b} & \checkmark & \checkmark & \checkmark & --\\
		Saveriano \etal\ \cite{Saveriano15} & -- & \checkmark & -- & \checkmark\\
		Towell \etal\ \cite{Towell10} & -- & \checkmark & -- & \checkmark\\
		Lin \etal\ \cite{Lin15} & \checkmark & \checkmark & -- & --\\
		Lober \etal\ \cite{Lober14,Lober15} & -- & \checkmark & -- & \checkmark\\
		Paraschos \etal\ \cite{Paraschos17} & -- & \checkmark & \checkmark & \checkmark\\
		\textbf{Our approach} & \checkmark & \checkmark & \checkmark & \checkmark\\
		\bottomrule[1.0pt]
	\end{tabular}
	\caption{Contributions of our approach with respect to previous works on task priority learning.}
	\label{tab:RelatedWork}
	\vspace{-0.0cm}
\end{table}

\subsubsection{Approaches related to soft weighting of tasks}
Concerning the soft weighting of tasks, control solutions are typically given by a combination of scalar-weighted sub-tasks, see \cite{Moro13} for an example with torque control and manually set weights. Dehio \etal\ \cite{Dehio15} and Modugno \etal\ \cite{Modugno16} propose to learn the weights of each sub-task using Covariance Matrix Adaptation Evolution Strategy (CMA-ES), a derivative-free stochastic optimization method. In \cite{Dehio15}, the weights are assumed to be constant throughout each execution, while in \cite{Modugno16}, they are parameterized by Radial Basis Functions spread along a time window, which allows the priorities of the different sub-tasks to change over time. Both approaches require the previous definition of a fitness function and a set of elementary tasks. Our approach also requires prior information about potential hierarchies, but, unlike \cite{Dehio15,Modugno16}, it directly exploits the demonstrations to discover the \iffalse structure of the task \fi employed prioritization, without requiring the defition of fitness functions. 

Lober \etal\ \cite{Lober14} use stochastic optimization to refine DMPs of incompatible tasks in order to render them compatible. 
%The pre-definition of information that can bias the optimization on how to disambiguate incompatibility (such as the desired duration of the movement of each arm's task) seems to be required for the method to succeed.
Their approach focuses on the re-organization of primitives in series, leaving out scenarios where they need to be executed in parallel with different levels of priority. 
In \cite{Lober15}, the approach was improved by exploiting Gaussian kernels to compute variance-dependent weights that are used to determine the priority of different sub-tasks. This approach shares connections with ours in that variance is used as a measure of task importance.
In that work, the variance depends on the distance to the kernel centers, which are pre-defined along the planned trajectory. 
In contrast, our method exploits the variability extracted from the demonstrations, allowing for learning new behaviors from the observed variations of a task. 

Another approach along the lines of the above works is that of Paraschos \etal\ \cite{Paraschos17}, who exploit the Probabilistic Movement Primitives (ProMP) framework to learn the accuracy of different tasks. This information, which reflects the variability of demonstrations given in Cartesian and joint spaces, is then used as a prioritization criteria to organize the different tasks. In this sense, that work shares stronger connections with \cite{Calinon09AR} than with our approach in Section \ref{sec:priorityConstraints} as it does not generate strict hierarchies to reproduce demonstrated movements. By identifying the hierarchies employed in the demonstrations, our approach ensures a more strict fulfillment of each individual task.% Additionally, ProMP-based techniques tend to require a high number of demonstrations in order to avoid numerical issues arising from singular parameter estimations.}

Table \ref{tab:RelatedWork} summarizes the main contributions of our approach for  learning task prioritization hierarchies, described in Section \ref{sec:priorityConstraints}, with respect to the state-of-the-art. The column \textit{Learn hierarchies} indicates whether or not the approach outputs strict hierarchies, as opposed to weights (scalar or matrix), while the column \textit{From demonstrations} distinguishes between LfD and RL approaches.

\section{Task-Parameterized Gaussian Mixture Models}
\label{sec:LearningTPMovements}

In previous work \cite{Calinon12Humanoids,Calinon16JIST} we introduced a probabilistic approach to the learning of task-parameterized movement primitives, with an example of implementation as Task-Parameterized Gaussian Mixture Model (TP-GMM). TP-GMM was used to encode demonstrated end-effector motions in multiple coordinate systems simultaneously, %described by a set of \textit{task parameters}
whose importance changed during the task depending on the variability observed in the demonstrations.
This information was exploited to adapt skills to new situations, in accordance to the demonstrated local features. In this section we review the TP-GMM formulation.

\subsection{Overview and nomenclature}

Throughout this article we will exploit extensively the linear properties of Gaussian distributions, which are central to TP-GMM. For this reason, in this subsection we review the concept of so-called \textit{task parameters} and give an overview of the technique.

\begin{definition}Task parameters are
sets of linear operators $ \mb{A}^{(j)},  \mb{b}^{(j)} $ that map Gaussian distributions from $ P $ \iffalse local feature \fi subspaces, indexed by ${j=1,\ldots,P}$, onto a common space. Such sets are called ``task parameters'' since they are part of the parameterization of a task, i.e., they influence how the robot accomplishes the given task.
\end{definition}

In TP-GMM, $ P $ \iffalse feature \fi subspaces encode local features of a demonstrated skill. 
%(Section \ref{sec:modelEst} describes how local models of features are trained). 
Once projected onto a common space, through \textit{task parameters}, the local models are combined to provide a solution fulfilling the most important features during task execution (Section \ref{sec:GMR}).
In previous work, task parameters have been used to represent poses of objects in a robot workspace, mapping local models of demonstrations (from the perspective of $ P $ different objects) onto a global coordinate system, typically the  robot base frame. In this case, $ \mb{A}^{(j)} $ is a rotation matrix \cite{Calinon12Humanoids,Calinon16JIST,Calinon14ICRA,Rozo16TRO} or a quaternion matrix \cite{Silverio15}, representing an object orientation, and $ \mb{b}^{(j)} $ is a translation vector, representing the origin of an object coordinate system with respect to the base frame of the robot. It was common in previous work to refer to task parameters as \textit{candidate frames} or \textit{candidate coordinate systems}. This is because, for a given task, each set of task parameters $ \mb{A}^{(j)}, \mb{b}^{(j)} $ may or may not influence the task execution, depending on the variability of the teacher's demonstrations in the corresponding coordinate system. Hence, each set $ \mb{A}^{(j)}, \mb{b}^{(j)} $ is considered to be a \textit{candidate} for affecting the task outcome.

In this work, we exploit the linear structure inherent to the task parameter formulation of TP-GMM to address the problems of: 1) simultaneously learning operational and configuration space constraints; and 2) learning priority hierarchies from demonstrations. We do this by taking a different perspective from past work where task parameters were candidate coordinate systems. We propose a formulation of task parameters that correspond to \textit{candidate projection operators} (Section \ref{sec:combiningConstraints}) and \textit{candidate task hierarchies} (Section \ref{sec:priorityConstraints}).

For clarity, Table \ref{tab:Nomenclature} summarizes the most important variables and symbols that are exploited throughout the article.

\begin{table}
	\centering
	\begin{tabular}{c | c  p{4.8cm}}
			& Variable / symbol & \hspace{1.5cm} Description \\
			\hline
			\multirow{15}{*}{\rot{TP-GMM}} &$ P $ & Number of task parameters \\
			&$ K $ & Number of Gaussian components \\
			&$\mb{A}^{(j)}$, $\mb{b}^{(j)}$ & Task parameters $ j \in [1, P] $\\
			&$\mb{\mu}^{(j)}_i$, $\mb{\Sigma}^{(j)}_i$ & Mean and covariance of Gaussian ${ i \in [1,K] }$ for task parameter $ j $\\
			&$\pi_i$ & Mixing coefficient of Gaussian $ i $\\
			&$ M $ & Number of demonstrations \\
			&$ T_m $ & Number of datapoints in demonstration $ m \in [1,M]$ \\
			&$ N $ & Total number of datapoints in a demonstration dataset \\
			&$ D $ & Dimension of datapoints \\
			&$ \mb{\xi}_t\in\mathbb{R}^D $ & Datapoint at time $ t $ \\
			&$ \mb{\xi}\in\mathbb{R}^{D\times N} $ & Demonstration dataset \\
			\hline
			\multirow{10}{*}{\rot{Robot}} & $ \mb{x}_L, \mb{x}_R \in\mathbb{R}^3 $ & Cartesian position of left and right end-effectors \\
			& $ \mb{\epsilon}_L, \mb{\epsilon}_R \in\mathcal{S}^3 $ & Orientation of left and right end-effectors as unit quaternions\\
			& $ \mb{R}_L, \mb{R}_R \in\mathrm{SO}(3) $ & Orientation of left and right end-effectors as rotation matrices\\
			& $ N_q $ & Number of robot joints \\
			& $ \mb{q}_L, \mb{q}_R \in\mathbb{R}^{N_q} $ & Joint angles of left and right arms\\
			& $ \mb{J}_L, \mb{J}_R \in\mathbb{R}^{6\times N_q} $ & Jacobian matrices of left and right arms\\
			& $ > $ & Priority operator (the task on the operator  left has \textit{priority over} the one on the right)\\
			\hline
	\end{tabular}
	\caption{ Summary of the notation.}
	\label{tab:Nomenclature}
	\vspace{-0.2cm}
\end{table}

\subsection{Model estimation}
\label{sec:modelEst}
%We consider tasks defined in both operational and configuration spaces, consisting on the fulfillment of constraints in either or both spaces (e.g. tracking desired end-effector poses or joint trajectories). During task execution, robot movements may largely depend on given goals, object poses or obstacles, which can be defined with coordinate systems. We call these variables \textit{task parameters} since they are part of the parameterization of a task, i.e. they influence how a robot accomplishes the given task goals. In TP-GMM, the knowledge of the task parameters is exploited to adapt a demonstrated skill to new situations  (e.g., unobserved positions and orientations of a manipulated object).

Each demonstration $m\in\{1,\ldots,M\}$ contains $T_m$ datapoints of dimension $ D $ forming a dataset of $N$ datapoints $\{\mb{\xi}_{t}\}_{t=1}^N$ with $N\!=\!\sum_{m=1}^{M}\!T_m$ and $ \mb{\xi}_{t}\in\mathbb{R}^D $.
$P$ task parameters, that map between subspaces $ j=1,\ldots,P $ and a common space, are defined at every time step $t$ by $\{\mb{A}^{(j)}_t,\mb{b}^{(j)}_t\}_{j=1}^P$. The demonstrations $\mb{\xi}\!\in\!\mathbb{R}^{D\times N}$ are observed from each subspace, forming $P$ local datasets ${\mb{X}^{(j)}\!\in\!\mathbb{R}^{D\times N}}$.\footnote{As an example, in previous work \cite{Calinon12Humanoids, Calinon16JIST,Rozo16TRO}, $ \mb{\xi}_t $ corresponded to end-effector positions, and local datasets could be computed with $ {\mb{X}^{(j)}_t = \mb{A}^{(j)^{-1}}_t \left(\mb{\xi}_t - \mb{b}^{(j)}_t\right) }$, with task parameters $\{\mb{A}^{(j)}_t,\mb{b}^{(j)}_t\}_{j=1}^P$ representing coordinate systems parameterized by the orientations and positions of $ P $ objects.}

The \textit{model parameters} of a TP-GMM with $K$ components are defined by $\big\{\pi_i,\{\mb{\mu}^{(j)}_i,\mb{\Sigma}^{(j)}_i\}_{j=1}^P\big\}_{i=1}^K$, where $\pi_i$ are the mixing coefficients and $\mb{\mu}^{(j)}_i$, $\mb{\Sigma}^{(j)}_i$ denote the center and covariance matrix of the $i$-th Gaussian \iffalse in candidate coordinate system $j$\fi in \iffalse feature\fi subspace $ j $.
Learning of the model parameters is achieved by log-likelihood maximization using an \emph{expectation-maximization} (EM) algorithm, see \cite{Calinon16JIST} for details. %subject to the constraint that the data in the different subspaces arose from the same source, resulting in an \emph{expectation-maximization} (EM) algorithm to iteratively update the model parameters until convergence, see \cite{Calinon16JIST} for details.

\subsection{Gaussian Mixture Regression}
\label{sec:GMR}

The learned model is used to reproduce movements in new situations. Each subspace $j$ encodes local features of the demonstrated movement. In new situations, i.e., for new values of the task parameters $\mb{A}^{(j)}_t, \mb{b}^{(j)}_t$, one needs to find a trade-off between each subspace solution. TP-GMM solves this problem by fusing the local models through projection in a common space, using the product of Gaussians. In this way, a new GMM with parameters $\{\pi_i,\mb{\hat{\mu}}_{t,i},\mb{\hat{\Sigma}}_{t,i}\}_{i=1}^K$ is automatically generated as
\begin{multline}
  \mathcal{N}\!\Big( \mb{\hat{\mu}}_{t,i} , \mb{\hat{\Sigma}}_{t,i} \Big)
  \;\propto\;
  \prod\limits_{j=1}^P \mathcal{N}\!\Big(
  \mb{\hat{\mu}}{}_{t,i}^{(j)} ,\; \mb{\hat{\Sigma}}{}_{t,i}^{(j)} \Big) ,
  \;\mathrm{with}\\
  \mb{\hat{\mu}}{}^{(j)}_{t,i} = \!\mb{A}^{(j)}_t \mb{\mu}^{(j)}_i \!+\! \mb{b}^{(j)}_t
  \;,\quad
  \mb{\hat{\Sigma}}{}^{(j)}_{t,i} = \!\mb{A}^{(j)}_t\mb{\Sigma}^{(j)}_i \mb{A}^{(j)^\trsp}_t ,
  \label{eq:TPGMM}
\end{multline}
where the Gaussian product is analytically given by
\begin{equation}
  \mb{\hat{\Sigma}}_{t,i} = \Big( \sum\limits_{j=1}^P {\mb{\hat{\Sigma}}{}^{(j)}_{t,i}}^{-1} \Big)^{-1}
  ,\quad
  \mb{\hat{\mu}}_{t,i} = \mb{\hat{\Sigma}}_{t,i} \sum\limits_{j=1}^P {\mb{\hat{\Sigma}}{}^{(j)}_{t,i}}^{-1}
  \mb{\hat{\mu}}{}^{(j)}_{t,i} .
  \label{eq:GaussProduct}
\end{equation}

%Note that $ \mb{\hat{\mu}}^{(j)}_{t,i}, \mb{\hat{\Sigma}}{}^{(j)}_{t,i} $ in \eqref{eq:TPGMM} map the local features, computed using \eqref{eq:TPprojection}, back to the original space (in robotics problems, usually the base of the robot), for  new $ \mb{b}^{(j)}_t, \mb{A}^{(j)}_t $, where Gaussian products are computed.
Equation \eqref{eq:TPGMM} maps local models onto a common space, where information from the different subspaces is fused according to Eq. \eqref{eq:GaussProduct}. Note that the task parameters $ \mb{A}^{(j)}_t , \mb{b}^{(j)}_t $ may vary during reproduction and take values different from those observed during demonstrations. In previous work \cite{Calinon12Humanoids, Silverio15, Calinon14ICRA,Rozo16TRO}, this property was exploited to adapt demonstrated skills to new situations (typically, new position and orientation of manipulated objects).

The obtained GMM is used to generate a reference trajectory distribution for the robot through Gaussian Mixture Regression (GMR). In this case, the datapoint $\mb{\xi}_t$ is decomposed into two subvectors $\mb{\xi}^\ty{I}_t$ and $\mb{\xi}^\ty{O}_t$, spanning the input and output dimensions of the regression problem, thus the GMM obtained from \eqref{eq:TPGMM} and \eqref{eq:GaussProduct} encodes the joint probability distribution ${\mathcal{P}(\mb{\xi}^{\ty{I}}_t,\mb{\xi}^{\ty{O}}_t) \sim \sum_{i=1}^K\pi_i\;\mathcal{N}\big(\mb{\hat{\mu}}_{t,i},\mb{\hat{\Sigma}}_{t,i}\big)}$. For a time-driven movement, $\mb{\xi}^\ty{I}_t$ corresponds to the current time step, while $\mb{\xi}^\ty{O}_t$ can be the end-effector pose or the joint angles of the robot. The task parameters $\mb{A}^{(j)}_t$ and $\mb{b}^{(j)}_t$ are also decomposed so that the input is not modulated by the task parameterization. Compared to an initial TP-GMM encoding $\mb{\xi}^\ty{O}_t$ with task parameters $\mb{A}^{\ty{O}(j)}_t$ and $\mb{b}^{\ty{O}(j)}_t$, the combination of TP-GMM and GMR instead encodes $\mb{\xi}_t\!=\!\big[{\mb{\xi}^\ty{I}_t}^\trsp \>\> {\mb{\xi}^\ty{O}_t}^\trsp \big]^\trsp$ with\footnote{in the case of a scalar time input, the identity matrix $\mb{I}$ collapses to $1$.} 
\begin{equation}
\mb{A}^{(j)}_t = \begin{bmatrix} \mb{I} & \mb{0} \\ \mb{0} & \mb{A}^{\ty{O}(j)}_t \end{bmatrix}
\;,\quad
\mb{b}^{(j)}_t = \begin{bmatrix} \mb{0} \\ \mb{b}^{\ty{O}(j)}_t \end{bmatrix}.
\end{equation}

Thus, GMR generates a new distribution ${\mathcal{P}(\mb{\xi}^{\ty{O}}_t | \mb{\xi}^{\ty{I}}_t) = \mathcal{N}\big(\mb{\xi}^{\ty{O}}_t | \mb{\hat{\mu}}^{\ty{O}}_t,\mb{\hat{\Sigma}}^{\ty{O}}_t\big)}$ that is used to control the robot.

\section{Learning operational and configuration space constraints simultaneously}
\label{sec:combiningConstraints}

%Previous works that exploit TP-GMM \cite{Silverio15,Calinon14ICRA,Rozo16TRO} consider manipulation problems defined in operational space. In such cases, task parameters are related to object poses, i.e., $ \mb{b}^{(j)}_t $ and $ \mb{A}^{(j)}_t $ parameterize positions and orientations of coordinate systems\iffalse, respectively\fi. However, robots are controlled in configuration space, i.e. even when tasks are defined in operational space, we map Cartesian velocity/force references into joint velocities/torques. As a consequence, it makes sense to also consider the configuration space when learning manipulation skills. Moreover, in learning problems, regardless of the algorithm or method that is employed to learn a skill, we typically choose either the configuration or the operational space to encode it, which requires prior reasoning about the task to be taught.

In this section, %we show that Jacobian matrices, commonly employed in robotics as linear mapping operators for differential kinematics, can be used as task parameters to simultaneously learn operational and configuration space constraints in a probabilistic way. 
we propose to exploit the structure of TP-GMM introduced in Section \ref{sec:LearningTPMovements} to simultaneously consider constraints in operational and configuration spaces. The aim is to circumvent the need for previously selecting the space in which one should encode a given task, and instead let the model automatically discover the proper space from a small set of demonstrations. The approach  consists of encoding the demonstrated movement in both spaces and, through statistics, extracting the space with the least variability (i.e., with the highest consistency) at each reproduction step. We do this by considering Jacobian-based task parameters (formulated in Sections \ref{sec:JacTP} and \ref{sec:JacTPorient}) that project operational space constraints onto configuration space (Section \ref{sec:JacTPcanonical}), where Gaussian products \eqref{eq:TPGMM} are computed.

Conceptually, this approach shares connections with the one introduced in \cite{Calinon09AR}. However, here we:
\begin{compactenum}
\item consider the learning of operational space constraints with respect to an arbitrary number of objects, by framing the approach in the context of TP-GMM, and
\item develop a formulation to consider end-effector orientation constraints.
\end{compactenum}
Additionally, the analysis focuses on humanoid robots and bimanual manipulation.

\subsection{Jacobian-based task parameters for position constraints}
\label{sec:JacTP}

Handling constraints in configuration and operational spaces is achieved by exploiting the task parameter linear structure in TP-GMM. Formally, consider a manipulator with $ N_q $ joints, whose positions and velocities are denoted by $ {\mb{q}, \mb{\dot{q}}\in\mathbb{R}^{N_q}} $. Its differential kinematics are given by $ \left[\mb{\dot{x}}^\trsp \>\> \mb{\omega}^\trsp \right]^\trsp = \mb{J}\mb{\dot{q}} $, where $ \mb{\dot{x}},\mb{\omega}\in\mathbb{R}^3 $ are the operational space linear and angular velocities. The Jacobian matrix $ \mb{J} = \left[\mb{J}^\trsp_p \>\> \mb{J}^\trsp_o\right]^\trsp \in \mathbb{R}^{6\times N_q}$ accounts for the contribution of joint velocities to operational space velocities, with matrices $\mb{J}_p,\mb{J}_o\in\mathbb{R}^{3\times N_q}$ responsible for the linear and angular parts, respectively. We here assume redundant manipulators, i.e. $ N_q > 6 $. 
%The inverse differential kinematics equation $ \mb{\dot{q}} = \mb{J}^\psin \mb{\dot{x}} $, with $ \mb{J}^\psin = \mb{J}^\trsp \left(\mb{J}\mb{J}^\trsp\right)^{-1} $ the right pseudo-inverse of the manipulator Jacobian $ \mb{J} $, yields the minimum-norm $ \mb{\dot{q}} $ that ensures $ \mb{\dot{x}} $ in operational space \cite{Siciliano09Book}.

The inverse differential kinematics for the position part of the operational space is given by $ \mb{\dot{q}} = \mb{J}_{\!p}^\psin \mb{\dot{x}} $, with $ {\mb{J}_{\!p}^\psin = \mb{J}_{\!p}^\trsp \left(\mb{J}_{\!p}\mb{J}_{\!p}^\trsp\right)^{-1}} $ the right pseudo-inverse of the Jacobian $ \mb{J}_{\!p} $. It yields the minimum-norm $ \mb{\dot{q}} $ that ensures $ \mb{\dot{x}} $ in operational space \cite{Siciliano09Book}. Numerical integration of this equation permits the computation of joint references for a desired end-effector position $ \mb{x}_t $ as (dropping the Jacobian subscript $ p $)
\begin{align}
\mb{\hat{q}}_t-\mb{q}_{t-1} & = \mb{J}_{t-1}^\psin \left(\mb{x}_t - \mb{x}_{t-1} \right) \nonumber \\
\iff \mb{\hat{q}}_t & = \mb{J}_{t-1}^\psin \mb{x}_t + \mb{q}_{t-1} - \mb{J}_{t-1}^\psin \mb{x}_{t-1},
\label{eq:JacTP1}
\end{align}
where $ \mb{\hat{q}}_t $ denotes the desired joint angles at $ t $. The structure of \eqref{eq:JacTP1}, being affine in $ \mb{x}_t $, allows us to connect inverse kinematics with TP-GMM. Let us assume, for the sake of the argument, that end-effector positions are modeled according to $ {\mb{x}_t\sim \sum^K_{i=1}\pi_i \mathcal{N}\left(\mb{\mu}^{(j)}_{i},\mb{\Sigma}^{(j)}_{i}\right)}$. The index $ j $ denotes one arbitrary subspace, as discussed in Section \ref{sec:LearningTPMovements}, where robot end-effector positions are locally modeled by a GMM. It follows that the linear transformation properties of Gaussian distributions \eqref{eq:TPGMM} can be applied to \eqref{eq:JacTP1} to project the local GMM onto the configuration space, resulting in 
\begin{equation}
	\mb{\hat{q}}^{(j)}_{t,i} = \underbrace{\mb{J}_{t-1}^\psin}_{\mb{A}^{(j)}_t} \mb{\mu}^{(j)}_{i} + \underbrace{\mb{q}_{t-1} - \mb{J}_{t-1}^\psin \mb{x}_{t-1}}_{\mb{b}^{(j)}_t}, \quad\forall i =1,\dots,K,
	\label{eq:JacTP2}
\end{equation}
for the center of the Gaussian $ i $, and
\begin{equation}
\mb{\hat{\Sigma}}^{(j)}_{t,i} = \underbrace{\mb{J}_{t-1}^\psin}_{\mb{A}^{(j)}_t} \mb{\Sigma}^{(j)}_{i} \underbrace{(\mb{J}_{t-1}^\psin)^\trsp}_{\mb{A}^{(j)^\trsp}_t}, \quad\forall i =1,\dots,K,
\label{eq:JacTP2_Sigma}
\end{equation}
for the corresponding covariance matrix. Equations \eqref{eq:JacTP2} and \eqref{eq:JacTP2_Sigma} show that Jacobian-based task parameters $ \mb{A}^{(j)}_t, \mb{b}^{(j)}_t $ act as \textit{projection operators} that map Gaussian distributions from operational to configuration space, creating distributions of joint angles. %The linear structure of TP-GMM is, hence, not exclusive to the use of coordinate system representations of task parameters, commonplace in previous works as we saw in Section \ref{sec:LearningTPMovements}.

This result is the cornerstone of more complex types of %operational space projection
operators. For instance, if we consider end-effector positions encoded with respect to an object parameterized by a translation vector $ \mb{p}^{(j)}_t $ and a rotation matrix $ \mb{R}^{(j)}_t $, \eqref{eq:JacTP2} becomes
\begin{equation}
\mb{\hat{q}}^{(j)}_{t,i} = \underbrace{\mb{J}_{t-1}^\psin \mb{R}^{(j)}_t}_{\mb{A}^{(j)}_t} \mb{\mu}^{(j)}_{i} + \underbrace{\mb{J}_{t-1}^\psin\left(\mb{p}^{(j)}_t - \mb{x}_{t-1}\right) + \mb{q}_{t-1}}_{\mb{b}^{(j)}_t},
\label{eq:JacTP4}
\end{equation}
which can be derived in a straightforward manner from \eqref{eq:JacTP2} by assuming rotated and translated Gaussians with center $ \mb{R}^{(j)}_t \mb{\mu}^{(j)}_{i} + \mb{p}^{(j)}_t $ and covariance $ \mb{R}^{(j)}_t \mb{\Sigma}^{(j)}_{i} \mb{R}^{(j)^\trsp}_t$. Similarly, the expression for the covariance matrix $ \mb{\hat{\Sigma}}^{(j)}_{t,i} $ can be easily obtained based on \eqref{eq:JacTP2_Sigma}, employing $ \mb{A}^{(j)}_t $ as defined in \eqref{eq:JacTP4}.

On the basis of this construction of task parameters, the local datasets $ \mb{X}^{(j)} = [\mb{X}^{(j)}_1,\ldots,\mb{X}^{(j)}_N] $ used to train the TP-GMM model can be constructed in a similar fashion as in Section \ref{sec:LearningTPMovements}. If a subspace $ j $ models the absolute end-effector position, i.e. with respect to the base frame of the robot, we have $\mb{X}^{(j)}_t = \mb{x}_t$. On the other hand, when $ j $ is associated with a coordinate system with pose parameters $ \mb{p}^{(j)}_t, \mb{R}^{(j)}_t $ as in \eqref{eq:JacTP4}, we have $ {\mb{X}^{(j)}_t = \mb{R}^{(j)^\trsp}_t\left(\mb{x}_t - \mb{p}^{(j)}_t\right) }$.

%$m\in\{1,\ldots,M\}$ contains $T_m$ datapoints of dimension $ D $ forming a dataset of $N$ datapoints $\{\mb{\xi}_{t}\}_{t=1}^N$ with $N\!=\!\sum_{m=1}^{M}\!T_m$ and $ \mb{\xi}_{t}\in\mathbb{R}^D $.
%$P$ task parameters, that map between subspaces $ j=1,\ldots,P $ and a common space}, are defined at every time step $t$ by $\{\mb{A}^{(j)}_t,\mb{b}^{(j)}_t\}_{j=1}^P$. The demonstrations $\mb{\xi}\!\in\!\mathbb{R}^{D\times N}$ are observed from \iffalse these different viewpoints\fi  \iffalse the feature \fi each subspace}, forming $P$ \iffalse trajectory samples\fi local datasets} ${\mb{X}^{(j)}\!\in\!\mathbb{R}^{D\times N}}$. 
%As an example, in previous work \cite{Calinon12Humanoids, Calinon16JIST,Rozo16TRO}, $ \mb{\xi}_t $ corresponded to end-effectors positions, and the task parameters represented coordinate systems, thus the local datasets were computed from $ {\mb{X}^{(j)}_t = \mb{A}^{(j)^{-1}}_t \left(\mb{\xi}_t - \mb{b}^{(j)}_t\right) }$ since $\{\mb{A}^{(j)}_t,\mb{b}^{(j)}_t\}_{j=1}^P$ parameterized the orientations and positions of $ P $ objects.}

In summary, in this novel formulation of task parameters, $ \mb{A}^{(j)}_t\in\mathbb{R}^{N_q\times3} $ and  $ \mb{b}^{(j)}_t\in\mathbb{R}^{N_q} $ map from Cartesian position to joint angles, solving the inverse kinematics with a reference given by the mean $ \mb{\mu}^{(j)}_i $. The TP-GMM representation is therefore extended to Jacobian-based, time-varying, task parameters $\{\mb{b}^{(j)}_t,\mb{A}^{(j)}_t\}_{j=1}^P$ with \iffalse different dimensionality for each candidate projection $j$ and \fi non-square $\mb{A}^{(j)}_t$ matrices. With this representation, operational space constraints are projected onto configuration space, where Gaussian products can be computed as in \eqref{eq:GaussProduct}, extending the original TP-GMM formulation to the consideration of configuration space constraints.

\begin{table*}
%	\begin{mdframed}
	\captionsetup{justification=centering}
	\setlength\tabcolsep{2.0pt} % default value: 6pt
	\renewcommand{\arraystretch}{2.0}
	\caption{Summary of task parameters as candidate projection operators}
	\centering
	\begin{tabular}{|p{5cm} l  l  l  l|}%[!t]
			\hline
			\quad \textit{Configuration space constraints:} &
%			&\begin{matrix} \text{\textit{Configuration space}}\\ \text{\textit{constraints:}}
%			\end{matrix}
			$ \mb{\hat{q}}^{(j)}_{t,i} $ & $ = \mb{I} $	& $ \mb{\mu}^{(j)}_i $ & $ + \> \mb{0}  $
%			\label{eqtab:canonicalFrame}
			\\
%			\hline
			\quad \textit{Absolute position constraints:} &
			%
%			&\begin{matrix} \text{\textit{Absolute position}} \\ \text{\textit{constraints:}}
%			\end{matrix}
			$ \mb{\hat{q}}^{(j)}_{t,i} $ & $ = \mb{J}^\psin_{t-1} $	& $ \mb{\mu}^{(j)}_i $ & $ - \> \mb{J}^\psin_{t-1} \mb{x}_{t-1} + \mb{q}_{t-1} $
%			\label{eq:Ab_xTable}
			\\
%			\hline
			\quad \textit{Relative position constraints:} &
			%
%			&\begin{matrix} \text{\textit{Relative position}} \\ \text{\textit{constraints:}}
%			\end{matrix}
			$ \mb{\hat{q}}^{(j)}_{t,i} $ & $ = \mb{J}^\psin_{t-1} \mb{R}^{(j)}_t $ & $ \mb{\mu}^{(j)}_i $ & $ + \> \mb{J}^\psin_{t-1} (\mb{p}^{(j)}_t - \mb{x}_{t-1}) + \mb{q}_{t-1} $
%			\label{eq:Ab_xobjTable}
			\\
%			\hline			
			\quad \textit{Absolute orientation constraints:} &
			%
%			&\begin{matrix} \text{\textit{Absolute orientation}} \\ \text{\textit{constraints:}}
%			\end{matrix}
			$ \mb{\hat{q}}^{(j)}_{t,i} $ & $ = \mb{J}^\psin_{t-1} \HRightVec\!(\mb{\bar{\epsilon}}_{t-1}) $ & $ \mb{\mu}^{(j)}_i $ & $ + \> \mb{q}_{t-1} $
%			\label{eq:Ab_quatTable}
			\\
%			\hline
			\quad \textit{Relative orientation constraints:} &
			%
%			&\begin{matrix} \text{\textit{Relative orientation}} \\ \text{\textit{constraints:}}
%			\end{matrix}
			$ \mb{\hat{q}}^{(j)}_{t,i} $ & $ = \underbrace{\mb{J}^\psin_{t-1} \HRightVec\!(\mb{\bar{\epsilon}}_{t-1}) \HLeft\!(\mb{\epsilon}^{(j)}_t)}_{\mb{A}^{(j)}_t} $ & $ \mb{\mu}^{(j)}_i $ & $ + \underbrace{\mb{q}_{t-1}\hspace{2.75cm}}_{\mb{b}^{(j)}_t} $\\
			\hline
%			\label{eq:Ab_quatobjTable}
	\end{tabular}
%	\end{mdframed}\\
	\label{tab:ProjOperators}	
\end{table*}

%\subsection{Orientation projection operators}
\subsection{Task parameters for orientation}
\label{sec:JacTPorient}

%So far we have considered the projection of position constraints on configuration space. However, as we saw for the case of coordination, orientation represents an important component of operational space constraints in most tasks.

%The previous subsection showed how we can formulate task parameters to project end-effector position constraints onto configuration space. 
Orientation constraints represent an important component of operational space in many tasks and in bimanual manipulation the orientation between end-effectors is of utmost importance for correct task execution \cite{Silverio15}. %Here, we introduce for the first time the projection operators for orientation constraints.
Here, we take advantage of the algebraic properties of unit quaternions to derive linear operators for projecting orientation constraints onto configuration space.

Let us consider the orientation part of the end-effector pose represented by a unit quaternion $ \mb{\epsilon} $ (Appendix \iffalse \ref{app:Quat}\fi A reviews this representation\footnote[3]{The appendices can be found as supplementary material at \href{http://joaosilverio.weebly.com/tro}{\texttt{http://joaosilverio.weebly.com/tro}}}) and the inverse differential kinematics for angular velocities, $ \mb{\dot{\mb{q}}} = \mb{J}^\psin_{\!o} \mb{\omega} $. From \cite{Pastor11} we have that
\begin{equation}
\mb{\omega}_t \approx \frac{\mathrm{vec}(\mb{\epsilon}_{t}*\mb{\bar{\epsilon}}_{t-1})}{\Delta t}
\label{eq:wFromVecPart}
\end{equation}
gives the angular velocity that rotates the unit quaternion $ \mb{\epsilon}_{t-1} $ into $ \mb{\epsilon}_t $, during $ \Delta t $. The operation $ \mathrm{vec}(\mb{\epsilon}_{t}*\mb{\bar{\epsilon}}_{t-1}) $ can be replaced by the matrix-vector product $ \HRightVec(\mb{\bar{\epsilon}}_{t-1})\>\mb{\epsilon}_t$ (see Appendix \iffalse\ref{app:TPorientation}\fi B), allowing us to write the inverse kinematics equation as (dropping the subscript \textit{o} in the Jacobian matrix)
\begin{align}
& \mb{\hat{\dot{\mb{q}}}}_t = \mb{J}^\psin_{t-1} \HRightVec(\mb{\bar{\epsilon}}_{t-1})\>\mb{\epsilon}_t \frac{1}{\Delta t}\>, \\
\iff & \mb{\hat{q}}_{t} = \mb{J}^\psin_{t-1} \HRightVec(\mb{\bar{\epsilon}}_{t-1})\>\mb{\epsilon}_t\> + \mb{q}_{t-1},
\label{eq:JacTPQuat1}
\end{align}
which has a similar structure to \eqref{eq:JacTP1}, being linear for the quaternion $ \mb{\epsilon}_t $. In a similar way as in Section \ref{sec:JacTP}, if $\mb{\mu}^{(j)}_i$ is the center of a Gaussian $ i $, encoding the absolute orientation of the end-effector in a subspace $j$, we take advantage of the structure in \eqref{eq:JacTPQuat1} to devise new task parameters $\mb{A}^{(j)}_t$, $\mb{b}^{(j)}_t$ that map a GMM from quaternion space to configuration space, namely
\begin{equation}
%\mb{\hat{q}}^{(j)}_{t,i} = \underbrace{\mb{J}_{\mb{\epsilon}}^{\psin}(\mb{q}) \frac{1}{2} \HLeft(\mb{\epsilon}_{t-1}) \VecOperatorMat \HRight(\mb{\bar{\epsilon}}_{t-1})}_{\mb{A}^{(j)}_t} \>\mb{\mu}^{(j)}_i + \underbrace{\mb{q}_{t-1}}_{\mb{b}^{(j)}_t}.\nonumber
\mb{\hat{q}}^{(j)}_{t,i} =
\underbrace{\mb{J}^{\psin}_{t-1} \HRightVec(\mb{\bar{\epsilon}}_{t-1})}_{\mb{A}^{(j)}_t} \>\mb{\mu}^{(j)}_i
+
\underbrace{\mb{q}_{t-1}}_{\mb{b}^{(j)}_t},
\label{eq:Ab_quat}
\end{equation}
for the center of Gaussian $ i $ in subspace $j$ (the covariance can be derived by following the same rules that we explained in Section \ref{sec:JacTP}).

For a desired end-effector orientation encoded in a coordinate system whose orientation is given by  $ \mb{\epsilon}^{(j)}_t $, \eqref{eq:Ab_quat} becomes
\begin{equation}
\mb{\hat{q}}^{(j)}_{t,i}  = \underbrace{\mb{J}^\psin_{t-1} \HRightVec\left(\mb{\bar{\epsilon}}_{t-1}\right) \>\HLeft\left(\mb{\epsilon}^{(j)}_t\right)}_{\mb{A}^{(j)}_t} \mb{\mu}^{(j)}_i + \underbrace{\mb{q}_{t-1}}_{\mb{b}^{(j)}_t},
\label{eq:JacTP5}
\end{equation}
where $ \HLeft $ is a quaternion matrix (see Appendix \iffalse \ref{app:Quat}\fi A).

For this formulation, if a subspace $ j $ models the absolute end-effector orientation, i.e. with respect to the base frame of the robot, we have $\mb{X}^{(j)}_t \!= \mb{\epsilon}_t$. When $ j $ is associated with a coordinate system with quaternion $ \mb{\epsilon}^{(j)}_t \!\!$ , then $ {\mb{X}^{(j)}_t = \HLeft\!\left(\mb{\bar{\epsilon}}^{(j)}_t\right)\!\mb{\epsilon}_t }$.

\begin{algorithm}[bt]
	\small
	\captionsetup{font=small}	
	\caption{Simultaneously learning constraints in operational and configuration spaces}
	\begin{algorithmic}[1]
		\Statex \hspace{-5mm}\textit{\textbf{Initialization}}
		\State Select candidate projection operators from Table \ref{tab:ProjOperators} based on the task at hand
		\begin{itemize}
			\item Canonical  operator $ \mb{A}^{(j)}_t \!=\! \mb{I}$, $\mb{b}^{(j)}_t \!=\! \mb{0}$, for encoding configuration space constraints
			\item Operational space operators, for absolute or relative position/orientation constraints in operational space
		\end{itemize}
		\State Collect demonstrations and compute the local datasets $ \mb{X}^{(j)} $ according to the chosen operators
	\end{algorithmic}
	\begin{algorithmic}[1]
		\Statex \hspace{-5mm}\textit{\textbf{Model training}}
		\State Apply EM \cite{Calinon16JIST} to obtain $\big\{\pi_i,\{\mb{\mu}^{(j)}_i,\mb{\Sigma}^{(j)}_i\}_{j=1}^P\big\}_{i=1}^K$ 
	\end{algorithmic}		
	\begin{algorithmic}[1]
		\Statex \hspace{-5mm}\textit{\textbf{Movement synthesis}}
		\For{$t=1,\dots,N$}
			\For{$j=1,\dots,P$}
				\State Update $ \{\!\mb{A}^{(j)}_t\!\!,\mb{b}^{(j)}_t\!\} $ according to Table \ref{tab:ProjOperators}
%				\Statex \hspace{0.95cm} $ \{\!\mb{A}^{(j)}_t\!\!,\mb{b}^{(j)}_t\!\} \!\!\leftarrow\!\! \{\!\mb{J}_{t-1},\mb{q}_{t-1},\mb{x}_{t-1},\mb{\epsilon}_{t-1},\mb{p}^{(j)}_{t}\!\!,\mb{R}^{(j)}_{t}\!\!,\mb{\epsilon}^{(j)}_{t}\!\} $
			\EndFor
			\For{$ i=1, \dots, K $}
			\State Compute $ \mb{\hat{\mu}}_{t,i} $ and $ \mb{\hat{\Sigma}}_{t,i} $ from \eqref{eq:TPGMM} and \eqref{eq:GaussProduct}
			\EndFor
			\State Apply GMR at $ \mb{\xi}^{\ty{I}}_t $:  ${\mathcal{P}(\mb{\xi}^{\ty{O}}_t | \mb{\xi}^{\ty{I}}_t) = \mathcal{N}\big(\mb{\xi}^{\ty{O}}_t | \mb{\hat{\mu}}^{\ty{O}}_t,\mb{\hat{\Sigma}}^{\ty{O}}_t\big)}$
			\State Use $ \mb{\hat{\mu}}^{\ty{O}}_t $ as joint references for the robot controller
			
		\EndFor
	\end{algorithmic}
	\label{alg:combiningConstraints}
\end{algorithm}

\subsection{Task parameters for configuration space constraints}
\label{sec:JacTPcanonical}

The previous two subsections provided task parameters that project operational space constraints onto configuration space. However, in order to consider both operational and configuration space constraints, one requires a local model of the configuration space demonstrations as well. Encoding configuration space movements in a TP-GMM is done using simple task parameters $ \mb{A}^{(j)}_t = \mb{I}$, $\>\mb{b}^{(j)}_t=\mb{0} $, corresponding to a canonical projection operator. In this case, the subspace is the configuration space itself, i.e. , $ \mb{\hat{q}}^{(j)}_{t,i} = \mb{\mu}^{(j)}_i $. The local datasets are thus computed from $ \mb{X}^{(j)}_t = \mb{q}_t $, where $ \mb{q}_t \in \mathbb{R}^{N_q}$ is the vector of robot joint angles at time step $ t $. \\

Table \ref{tab:ProjOperators} gives a summary of the operators derived in this section. The overall procedure for learning and reproducing a skill is summarized in Algorithm \ref{alg:combiningConstraints}.

\subsection{Experiment: bimanual shaking skill}
\label{subsec:shakingTask}

\begin{figure}
	\centering
	\includegraphics[width=\columnwidth]{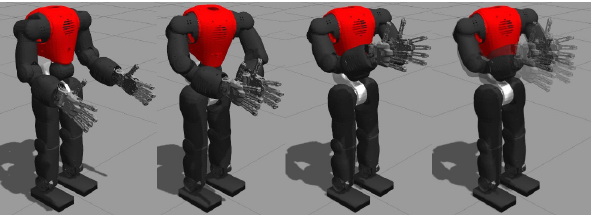}
	\caption{The COMAN robot performs the bimanual shaking task in simulation. \textbf{First:} The robot is in a neutral starting pose. \textbf{Second:} Reaching for the bottle and grasping it. \textbf{Third:} Bringing the bottle close to the torso. \textbf{Fourth:} Shaking movement executed through rhythmic oscillations of both shoulder joints.}
%	\vspace{-0.2cm}
	\label{fig:shakingCOMAN_Sim}
\end{figure}

In order to test the formulation introduced in this section, we selected the skill of shaking a bottle using COMAN. The skill contains an operational space component (reaching, grasping a bottle and bringing it closer to the torso) and a configuration space component (shaking with rhythmic shoulder movements, see Fig.\ \ref{fig:shakingCOMAN_Sim}). Parts of this experiment are also reported in \cite{Silverio2018Humanoids}.	%Such skill combines operational and configuration space constraints: on one hand there are constraints related to the pose of the object to be grasped, bimanual pose constraints, to keep relative poses constant, and a rhythmic joint movement in configuration space that performs the act of shaking, regardless of the end-effector pose in operational space. 

The upper-body of the COMAN robot comprises 17 DOFs: 3 DOFs for the waist and 7 for each arm, with the kinematic chains of both arms sharing the 3 waist joints. We define the differential kinematics of the left and right end-effectors as $\left[\mb{\dot{x}}^\trsp_L \>\> \mb{\omega}^\trsp_L \>\> \mb{\dot{x}}^\trsp_R \>\> \mb{\omega}^\trsp_R\right]^\trsp = \mb{J}_{\!\text{up}}\mb{\dot{q}}$, where $ \mb{J}_{\!\text{up}} $ is the upper-body Jacobian, $\mb{\dot{x}}_L, \mb{\omega}_L$, $\mb{\dot{x}}_R, \mb{\omega}_R$ are the left and right end-effector velocities and $\mb{\dot{q}}=\left[\mb{\dot{q}}^\trsp_{W} \>\> \mb{\dot{q}}^\trsp_L \>\> \mb{\dot{q}}^\trsp_R\right]^\trsp$ represents the concatenation of waist, left and right arm joint velocities. The Jacobian matrix is given by \cite{Lee14}
\begin{equation}
\mb{J}_{\!\text{up}} = \left[\begin{matrix}	\mb{J}_{W|L} & \mb{J}_{\!\!L} & \mb{0} \\
\mb{J}_{W|R} & \mb{0}   & \mb{J}_{\!\!R}     \end{matrix}\right],
\end{equation}
where $\mb{J}_{W|L}$, $\mb{J}_{W|R}$ denote the Jacobians that account for the effect of the waist joints on left and right end-effector velocities. $\mb{J}_{\!\!L}$ and $\mb{J}_{\!\!R}$ correspond to the Jacobians of the left and right end-effectors from the waist link. The inverse kinematics solution is given in this case by ${\mb{\hat{\dot{q}}} = \left[\begin{matrix} \mb{\hat{\dot{q}}}^\trsp_{W} \> \mb{\hat{\dot{q}}}^\trsp_L \> \mb{\hat{\dot{q}}}^\trsp_R \end{matrix}\right]^\trsp \!\!=\! \mb{J}^\psin_{\!\text{up}}\> \left[\begin{matrix} \mb{\dot{x}}^\trsp_L \> \mb{\omega}^\trsp_L \> \mb{\dot{x}}^\trsp_R \> \mb{\omega}^\trsp_R\end{matrix}\right]^\trsp}\!\!$, where $\mb{J}^\psin_{\!\text{up}}$ is the right pseudo-inverse of $ \mb{J}_{\!\text{up}} $.

\begin{figure}
	\begin{subfigure}{\columnwidth}
		\centering
		\includegraphics[width=\columnwidth]{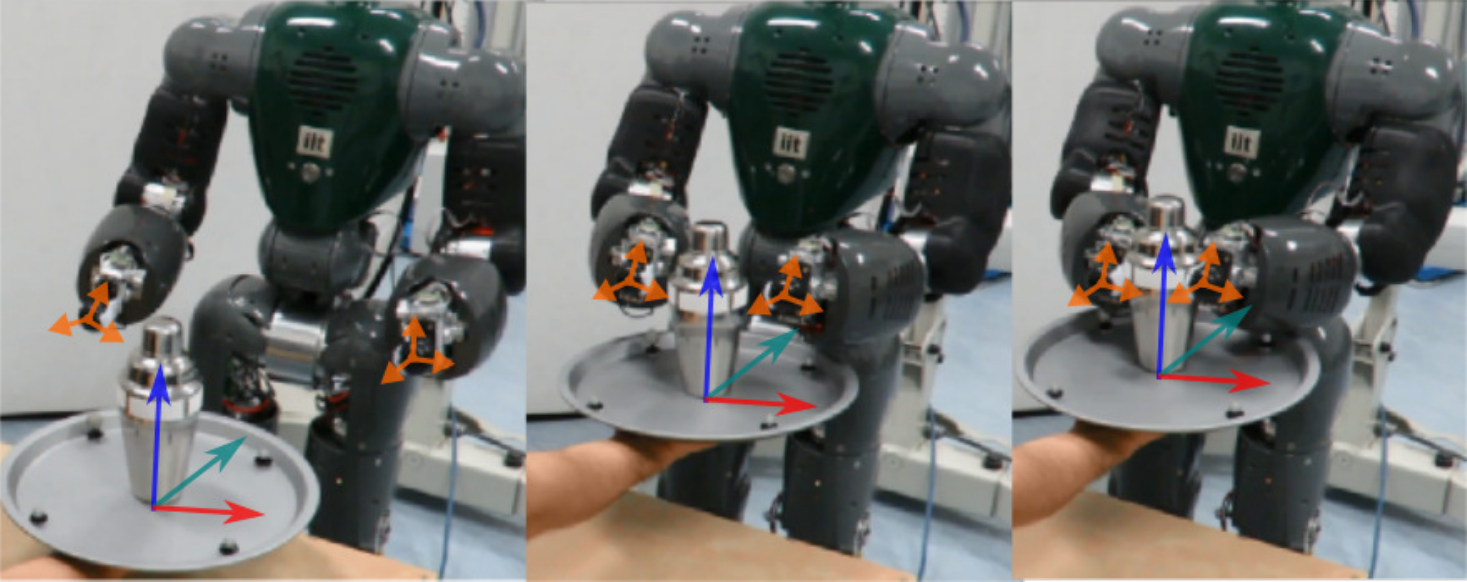}
		\vspace{-0.45cm}
	\end{subfigure}
	\begin{subfigure}{\columnwidth}
		\centering
		\includegraphics[width=\columnwidth]{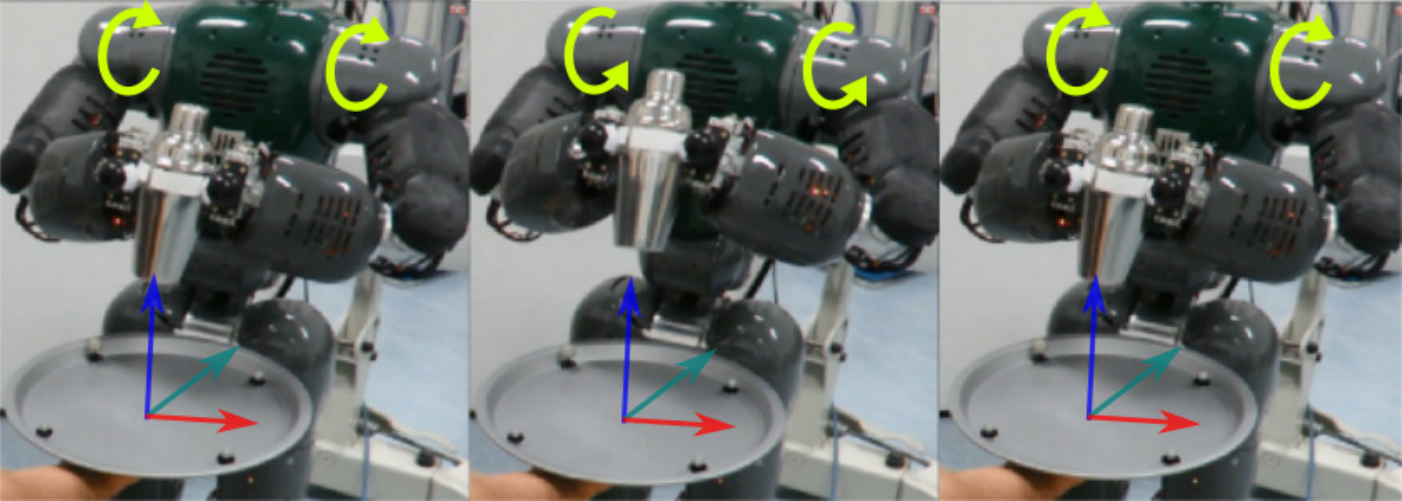}
	\end{subfigure}
	\caption{Reproduction of the shaking task in the real COMAN robot. \textbf{Top:} Snapshots of the reaching part of the movement, defined by constraints in operational space. The robot gradually reaches for the object, which is tracked using an optical system. \textbf{Bottom:} Shaking part of the movement, defined by constraints in configuration space. The robot performs the shaking through rhythmic motions of the shoulders, moving the shaker up and down repeatedly.}
	\label{fig:shakingCOMAN}
%	\vspace{-0.1cm}
\end{figure}

The experiment was conducted in the Gazebo simulator, and the skill was reproduced in the real robot (Figure \ref{fig:shakingCOMAN}). The demonstrations were generated by solving inverse kinematics in Gazebo, for the operational space part of the movement, and using sinusoidal references to control the shoulder joints for the shaking part\footnote{Alternatively, kinesthetic teaching or optical tracking of movements from humans could be used.}. Here, we make the assumption that the demonstrated grasp is always successful and the object will move together with the end-effectors after it is grasped. Hence, the pose of the bottle that is considered in this experiment is the one at the beginning of each demonstration. We also assume that the grasping points on the bottle are the same in all demonstrations.
We collected 10 demonstrations of the skill, each of them with different initial bottle poses and a duration of approximately 13 seconds. In each demonstration, we recorded both the joint angles and the end-effector poses with respect to the initial bottle frame. Temporal alignment of the demonstrations was achieved using Dynamic Time Warping \cite{Chiu04}. We used a TP-GMM with $K=10$ components, with $P=2$ projection operators ($K$ and $P$ chosen empirically). The first operator is a concatenation of \eqref{eq:JacTP4}, required for considering end-effector positions, and \eqref{eq:JacTP5}, for orientations, parameterized with the bottle pose $ \{\mb{p}^{(1)}_t \!\!, \mb{R}^{(1)}_t \!\!, \mb{\epsilon}^{(1)}_t\} $,
\setlength{\arraycolsep}{1pt}
%\begin{align}
%	&\mb{A}^{(1)}_t = \nonumber\\ &\mb{J}^\psin_{\!\text{up}} \!\!\left[\begin{matrix} \mb{R}^{(1)}_t & \mb{0} & \mb{0} & \mb{0}\\	
%																			  \mb{0} & \HRightVec(\mb{\bar{\epsilon}}_{L, t-1}) \HLeft(\mb{\epsilon}^{(1)}_t) & \mb{0} & \mb{0} \\
%																			  \mb{0} & \mb{0} & \mb{R}^{(1)}_t & \mb{0}\\	
%  																			  \mb{0} & \mb{0} & \mb{0} & \HRightVec(\mb{\bar{\epsilon}}_{R, t-1}) \HLeft(\mb{\epsilon}^{(1)}_t)\\
%																			  \end{matrix} \right]\!\!,\\
%	&\hspace{1.5cm}\mb{b}^{(1)}_t = \mb{J}^\psin_{\text{up}} \left[\begin{matrix} \mb{p}^{(1)}_t - \mb{x}_{L,t-1} \\ \mb{0} \\ \mb{p}^{(1)}_t - \mb{x}_{R,t-1} \\ \mb{0} \end{matrix}\right] + \mb{q}_{t-1}.
%\end{align}
\begin{align}
&\!\!\!\mb{A}^{(1)}_t \!\!=\! \mb{J}^\psin_{\!\text{up}} \!\!\left[\begin{smallmatrix} \mb{R}^{(1)}_t & \mb{0} & \mb{0} & \mb{0}\\	
\mb{0} & \!\!\!\HRightVec(\mb{\bar{\epsilon}}_{L, t-1}) \HLeft(\mb{\epsilon}^{(1)}_t) & \mb{0} & \mb{0} \\
\mb{0} & \mb{0} & \!\!\mb{R}^{(1)}_t & \mb{0}\\	
\mb{0} & \mb{0} & \mb{0} & \!\!\HRightVec(\mb{\bar{\epsilon}}_{R, t-1}) \HLeft(\mb{\epsilon}^{(1)}_t)\\
\end{smallmatrix} \right]\!\!,\!\!\!\\
&\hspace{1.5cm}\mb{b}^{(1)}_t = \mb{J}^\psin_{\text{up}} \left[\begin{smallmatrix} \mb{p}^{(1)}_t - \mb{x}_{L,t-1} \\ \mb{0} \\ \mb{p}^{(1)}_t - \mb{x}_{R,t-1} \\ \mb{0} \end{smallmatrix}\right] + \mb{q}_{t-1}.
\end{align}
For these task parameters, training datapoints are concatenations of both end-effector poses from the perspective of the bottle frame, i.e. %${ \mb{X}^{(1)}_t = \left[\mb{R}^{(1)^\trsp}_t\!\!\left(\mb{x}^\trsp_{L,t}- \mb{p}^{(1)}_t\right)\>\> \HLeft\!\left(\mb{\bar{\epsilon}}^{(1)}_t\right)\!\mb{\epsilon}^\trsp_{L,t} \>\> \mb{R}^{(1)^\trsp}_t\!\!\left(\mb{x}^\trsp_{R,t}- \mb{p}^{(1)}_t\right) \>\> \HLeft\!\left(\mb{\bar{\epsilon}}^{(1)}_t\right)\!\mb{\epsilon}^\trsp_{R,t}\right]^\trsp}$. The second projection operator is canonical, ${\mb{A}^{(2)}_t =  \mb{I}}, \> {\mb{b}^{(2)}_t = \mb{0}}$.
\begin{equation}
\mb{X}^{(1)}_t = \left[\begin{smallmatrix}
	\mb{R}^{(1)^\trsp}_t\!\!\left(\mb{x}^\trsp_{L,t}- \mb{p}^{(1)}_t\right)\\ \HLeft\!\left(\mb{\bar{\epsilon}}^{(1)}_t\right)\!\mb{\epsilon}^\trsp_{L,t} \\ \mb{R}^{(1)^\trsp}_t\!\!\left(\mb{x}^\trsp_{R,t}- \mb{p}^{(1)}_t\right) \\ \HLeft\!\left(\mb{\bar{\epsilon}}^{(1)}_t\right)\!\mb{\epsilon}^\trsp_{R,t}
\end{smallmatrix}\right].
\end{equation}
The second projection operator is canonical, ${\mb{A}^{(2)}_t =  \mb{I}}, \> {\mb{b}^{(2)}_t = \mb{0}}$.
In the above, subscripts $ L $ and $ R $ denote left and right end-effectors.
\begin{figure}[h]
	\centering
	\includegraphics[width=0.9\columnwidth]{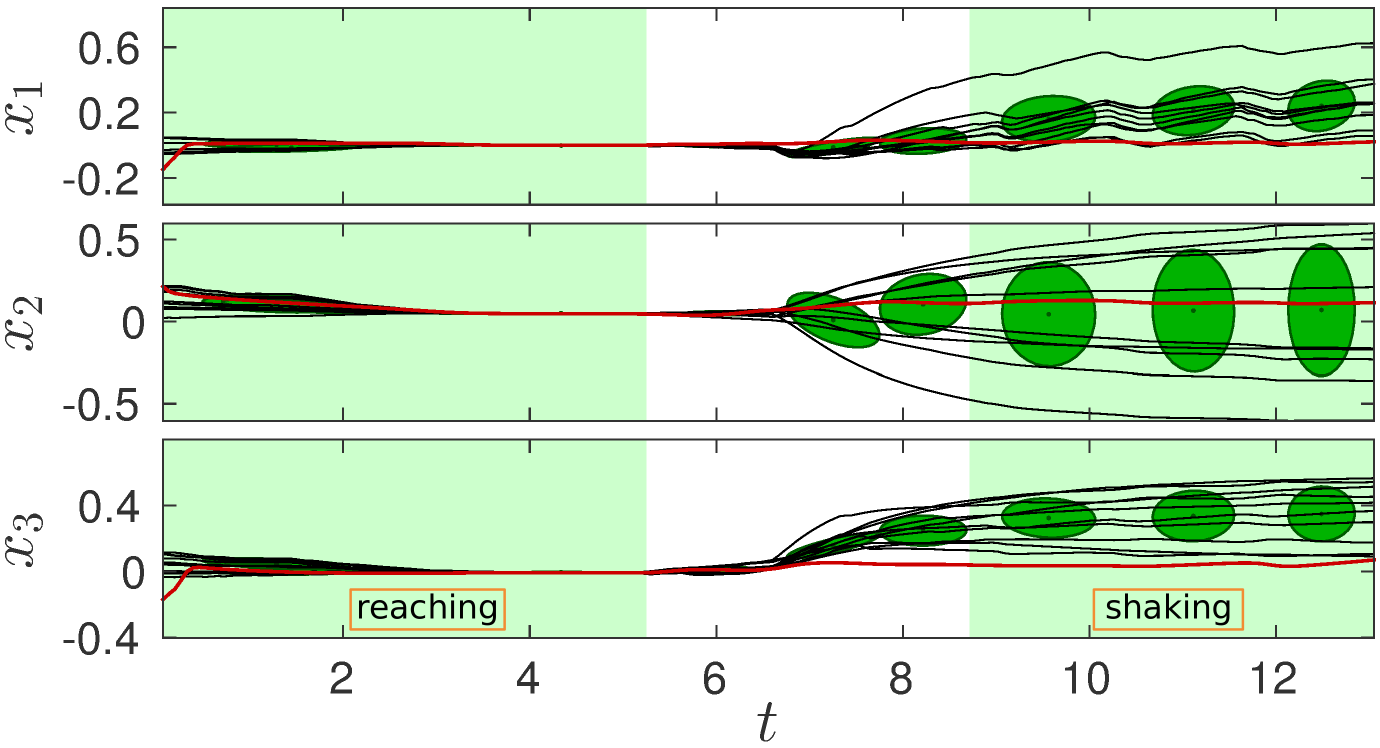}
	\vspace{-0.1cm}
	\caption{Left end-effector position (in meters) \iffalse from the perspective of the bottle coordinate system \fi with respect to the initial bottle coordinate system (prior to the grasp). Black lines represent demonstrations while the red line represents one reproduction of the movement. Ellipses depict the Gaussian components of the model (isocontour of one standard deviation). The shaded areas mark the duration of the reaching and shaking phases. Notice the low variability in position at the end of the reaching movement.}
	\label{fig:endEffectorPosShaking}
\end{figure}
\begin{figure}[h]
	\centering
	\includegraphics[width=0.9\columnwidth]{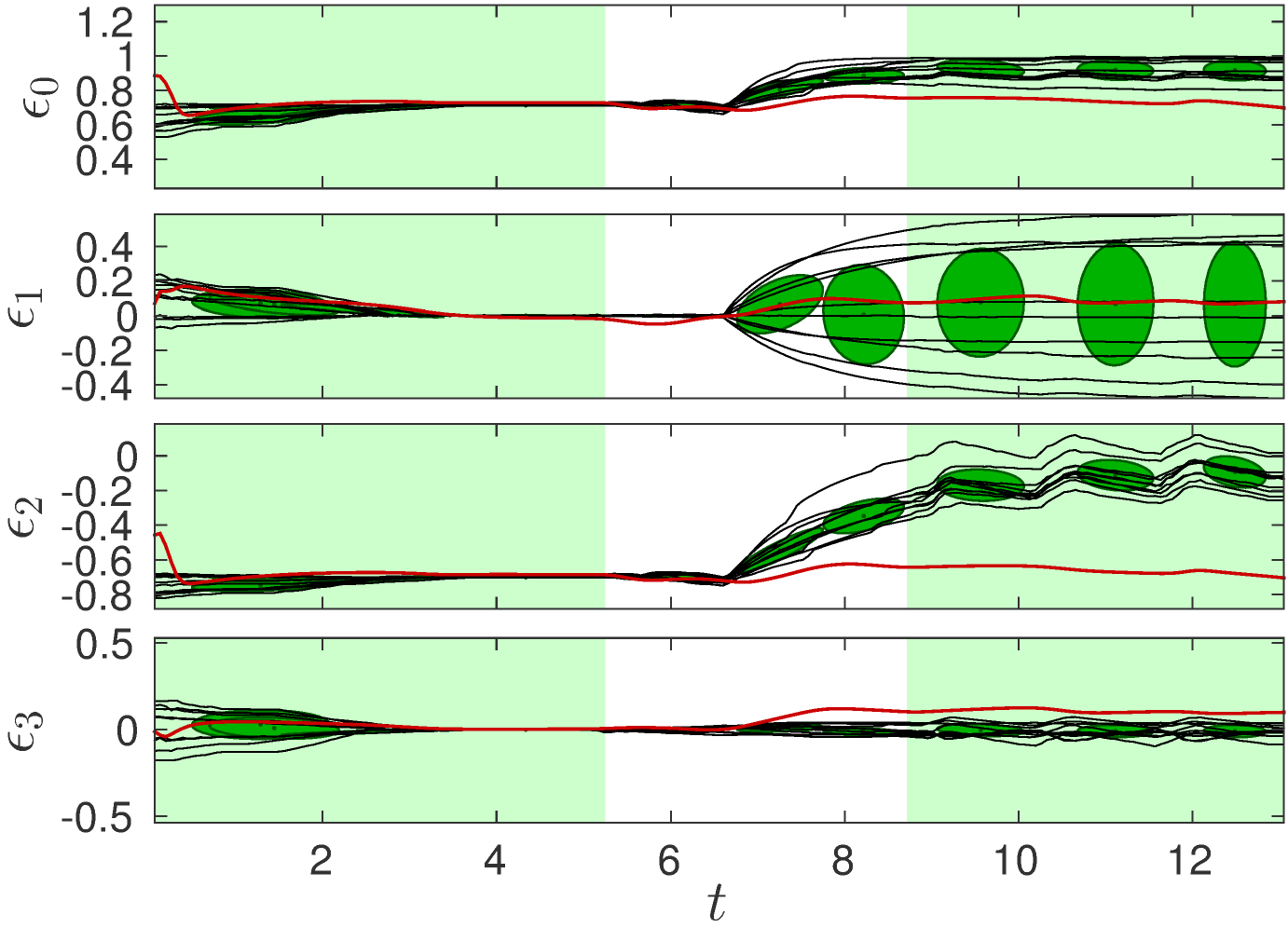}
	\vspace{-0.1cm}
	\caption{Left end-effector orientation (as a unit quaternion) \iffalse from the perspective of the bottle coordinate system \fi with respect to the initial bottle coordinate system (prior to the grasp). Notice the low variability in orientation at the end of the reaching movement.}
%	\vspace{-0.2cm}
	\label{fig:endEffectorOrientShaking}
\end{figure}
\begin{figure}[h]
	\centering
	\includegraphics[width=0.9\columnwidth]{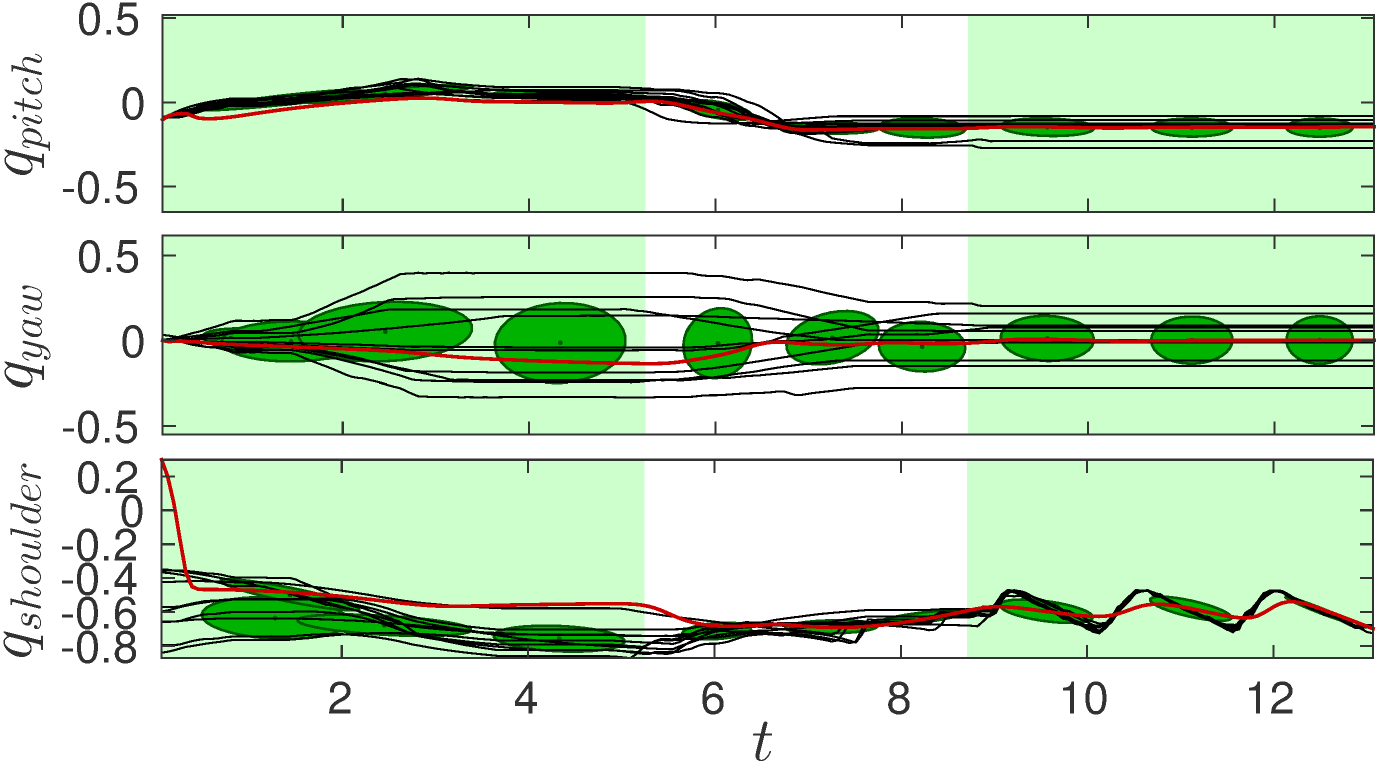}
	\vspace{-0.1cm}
	\caption{Waist joints (pitch and yaw) and shoulder joint of the left arm (in radians). Notice the low variability of the shoulder joint during the shaking part of the movement (second shaded area) and how the model captures the shaking pattern encoded in this joint.}
	\label{fig:JointsShaking}
\end{figure}

Figures \ref{fig:endEffectorPosShaking}--\ref{fig:JointsShaking} show the demonstration data over time\footnote{Due to space limitations we only plot data corresponding to the left arm.} (black lines), in operational and configuration spaces, together with the Gaussian components obtained after EM (green ellipses). In addition we also plot the references generated by GMR (red lines), for a new position and orientation of the bottle. In Figures \ref{fig:endEffectorPosShaking} and \ref{fig:endEffectorOrientShaking} we see that, during the reach and grasp movement, there is low variability in the demonstrations, both in position and orientation, when the end-effector is touching the bottle ($ t \approx 4s $). This is successfully encoded by the model (narrow Gaussians showing low variance), as this aspect of the skill is important for a correct completion of the task. It follows that the synthesized movement (red line) closely matches the demonstrations in the regions of low variability. Note that, after the grasp ($ t>7s $), the variance increases as the end-effectors move away from the initial bottle pose to perform the shaking movement. Figure \ref{fig:JointsShaking} shows that, from the beginning of the shaking phase ($ t\approx8s $), the shoulder joint (bottom graph) exhibits a consistent oscillatory pattern modeled by 3 Gaussians, which is adequately captured and synthesized by the model. This contrasts with the other joints of the robot, which do not influence the shaking. 

\begin{figure}
	\centering
	\includegraphics[width=0.8\columnwidth, trim={0cm 8.25cm 0cm 9.25cm},clip]{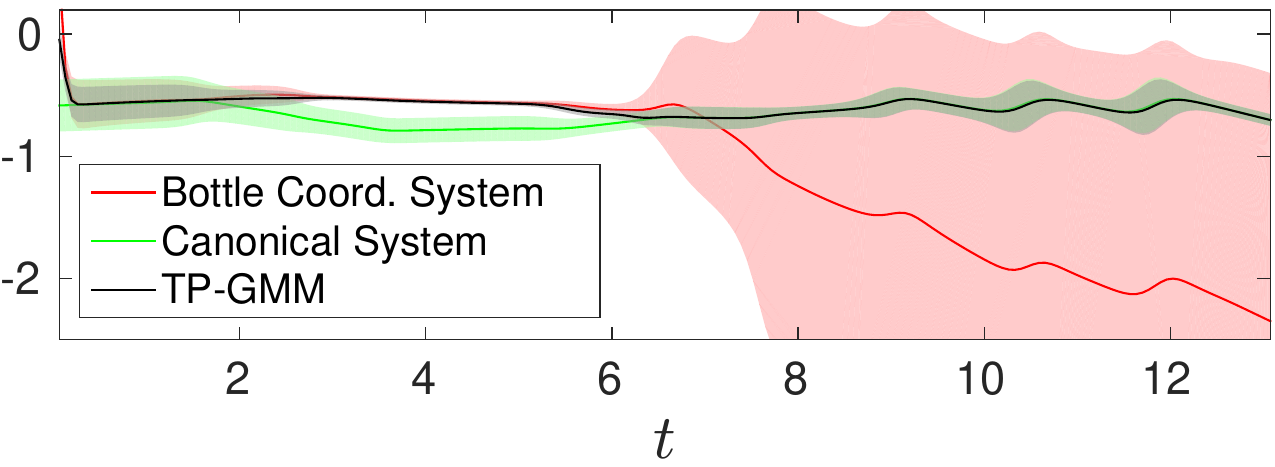}
	\vspace{-0.75cm}
	\caption{Shoulder joint angle estimation (radians) from each space (red and green) and the resulting reference after TP-GMM (black). Each estimate has an associated mean (solid line) and variance (light color envelope), learned from demonstrations and synthesized during reproduction.}
	\label{fig:ShoulderJointTPGMM}
%	\vspace{-0.0cm}
\end{figure}

When using the task parameters defined in this section, each set of operators is responsible for a candidate configuration space solution. The weight of each solution is estimated from the demonstrations, based on the variability in the data, and encoded by the different Gaussians as full covariance matrices. In this sense, the approach implements a form of task prioritization based on a \textit{fusion of tasks} (see Section \ref{sec:relatedWork} for a review of prioritization strategies), where full precision matrices act as weights on the candidate solutions. Figure \ref{fig:ShoulderJointTPGMM} shows that TP-GMM correctly extracted the most relevant configuration space solution according to the requirements of the overall task. Notice how, until $t\!\approx\!5s$, TP-GMM (black line) matches the candidate solution given by the bottle coordinate system (red line). Since the variability encoded in this coordinate system is low compared to that of the configuration space (because reaching and grasping is done in operational space), the Gaussian product favors this solution. This is achieved through the linear transformation properties of Gaussians, that allow for both centers and covariance matrices to be locally mapped from operational to configuration space using the proposed linear operators. Similarly, during the shaking phase, after $t\!\approx\!8s$, the reference generated for the shoulder joint matches the solution obtained using the canonical projection operator. This is because the shaking movement results in a consistent oscillatory pattern of shoulder joints, observed during the demonstrations, as seen in Figure \ref{fig:JointsShaking}. These results show that the proposed TP-GMM formulation is a viable solution for encoding task relevant features in both configuration and operational spaces, including orientation.

Finally, Figure \ref{fig:shakingCOMAN} shows the two distinct phases of the movement during a reproduction in the real COMAN robot. In this experiment, we used a tray to carry a shaker towards COMAN, where an optical tracking system provided the shaker pose to the robot. We employed the learned TP-GMM to generate joint references %$ \mb{\hat{q}}_t $
at every time step of the reproduction, which were fed to a joint position controller. In the top row of Figure \ref{fig:shakingCOMAN}, the robot takes into account the operational space constraints as it reaches for the bottle, while in the bottom row, the robot shakes the grasped bottle with rhythmic shoulder movements. In both parts of the movement, the operational and configuration space constraints are properly replicated. Videos are available at \mbox{\texttt{\footnotesize\url{http://joaosilverio.weebly.com/tro}} }.

\section{Learning task prioritization hierarchies from demonstrations}
\label{sec:priorityConstraints}

Controlling robots often requires the definition of priorities between tasks. For example, when a humanoid is standing and has to manipulate an object, the highest priority should be on keeping balance and, therefore, all degrees of freedom should be assigned to that task and only manipulate when balance is not compromised. %Another problem that frequently arises in humanoids is that, since the torso joints are shared by the kinematic chain of the two arms, simultaneously reaching for objects on the left and right sides of the robot yields poor results when the robot does not know which side to prioritize.
Commonly, the way in which tasks are prioritized is defined beforehand by an expert \cite{Moro13,Sentis05}. In contrast, here we propose to learn these priorities from demonstrations and develop a framework to do so.
We frame the problem in the context of TP-GMM by formulating the task parameters as candidate hierarchies %--- as opposed to candidate coordinate systems in the original formulation ---
and subsequently learning how to fuse the different hierarchies from demonstrations. Our approach is hence comprised of two components:
\begin{enumerate}
	\item \textbf{Hierarchy identification (Section \ref{subsec:extractingHierarchies}):} From a set of candidate task hierarchies, and knowledge of the possible control references for each sub-task, we identify the employed hierarchies based on the variability present in the demonstrations. 
	\item \textbf{Movement synthesis (Section \ref{subsec:softWeighting}):} A controller is used to reproduce the taught priority behaviors in new situations, through a fusion of candidate hierarchies.
\end{enumerate}

\begin{figure}
	\begin{subfigure}{\columnwidth}
		\centering
		\includegraphics[width=\columnwidth]{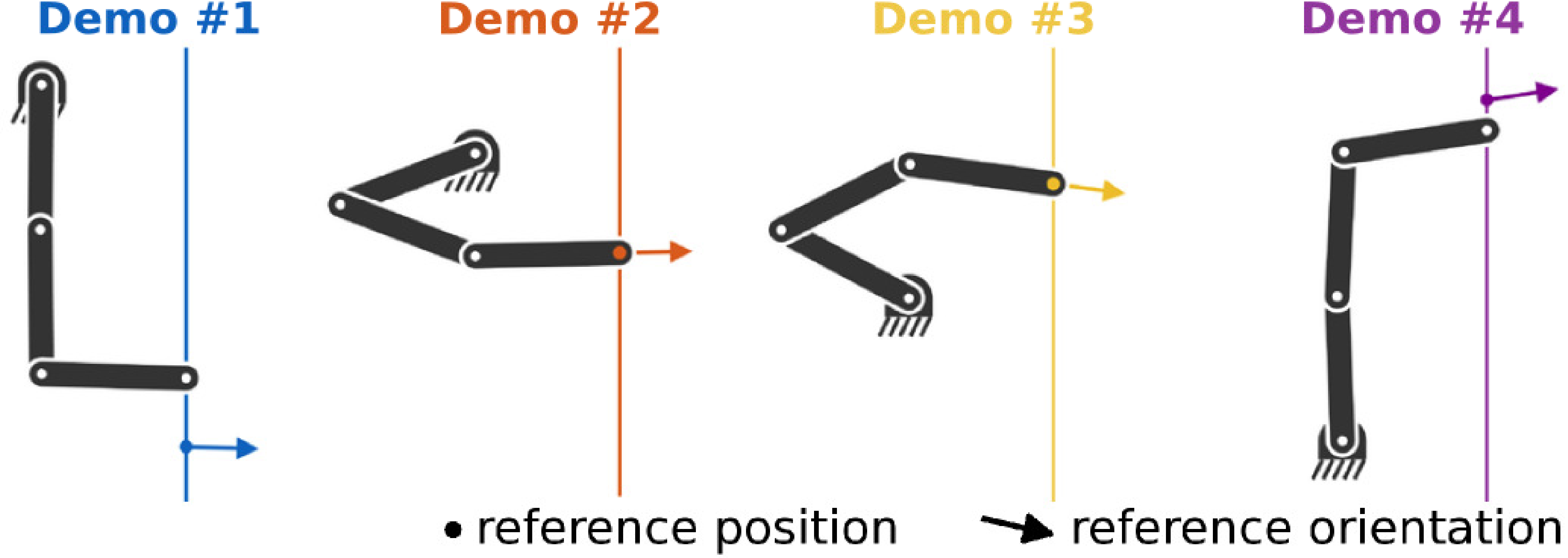}
		\caption{Four demonstrations of priority on \textbf{orientation}. The robot only fulfills the two tasks when they are both achievable (demonstrations 2 and 3). When they are incompatible, orientation is prioritized (demonstrations 1 and 4).}
		\vspace{0.5cm}
		\label{fig:planarPrioritiesOrient}
	\end{subfigure}
	\begin{subfigure}{\columnwidth}
		\centering
		\includegraphics[width=\columnwidth]{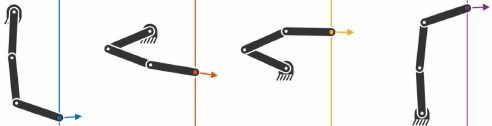}
		\caption{Four demonstrations of priority on \textbf{position}. Once the two tasks become incompatible the robot prioritizes position tracking (demonstrations 1 and 4).}
		\vspace{0.25cm}
		\label{fig:planarPrioritiesPos}
	\end{subfigure}
	\caption{Planar robot tracking an orientation (arrow) and a vertical position (dot) along a line, with different priorities.}
	\label{fig:planarPriorities}
	\vspace{-0.2cm}
\end{figure}

\subsection{Identifying priority hierarchies from demonstrations}\label{subsec:extractingHierarchies}

We consider %prioritization primitives
joint velocity commands generated from strict hierarchies, where tasks of lower importance are performed in the null space of more important ones, i.e., they are only executed if they do not conflict. In this subsection we show how %to extract the activation of the prioritization primitives
the prioritization employed during demonstrations can be identified from a set of candidate hierarchies.

In order to ease the explanation, let us consider an example of a 3-DOF planar robot with position and orientation tracking tasks (Figure \ref{fig:planarPriorities}). We denote the operational space velocities that ensure the tracking of position and orientation references by $\mb{\dot{x}}_1 = \mb{K}_p\mb{e}_p$ and $\mb{\dot{x}}_2 = \mb{K}_o\mb{e}_o$, respectively, where $ \mb{K}_p, \mb{K}_o $ are positive gains and $ \mb{e}_p = \mb{\hat{x}}_p - \mb{x}_p, \mb{e}_o = \mb{\hat{x}}_o - \mb{x}_o$ are the task errors (in principle, any controller that generates a command proportional to the error in the reference state would be a valid choice). These two tasks can be prioritized according to
\begin{equation}
\mb{\hat{\dot{\mb{q}}}}^{(1)}
= \mb{J}_{\!1}^\psin \; \mb{\dot{x}}_1 + \mb{N}_{\!1} \mb{J}_{\!2}^\psin \; \mb{\dot{x}}_2 \\
%&= \mb{J}_1^\trsp {(\mb{J}_1\mb{J}_1^\trsp)}^{-1} \; \mb{\dot{x}}_1 + 
%\left(\mb{I} - \mb{J}_1^\trsp {(\mb{J}_1\mb{J}_1^\trsp)}^{-1} \mb{J}_1\right) \; 
%\mb{J}_2^\trsp{(\mb{J}_2\mb{J}_2^\trsp)}^{-1} \; \mb{\dot{x}}_2 \\
= \underbrace{\left[\mb{J}_{\!1}^\psin \quad \mb{N}_{\!1}\mb{J}_{\!2}^\psin \right]}_{\mb{A}^{(1)}} 
\left[\begin{matrix} \mb{\dot{x}}_1 \\ \mb{\dot{x}}_2 \end{matrix}\right]
\label{eq:strictPriorityx1}
\end{equation}
or, alternatively,
\begin{equation}
\mb{\hat{\dot{\mb{q}}}}^{(2)} 
= \mb{J}_{\!2}^\psin \; \mb{\dot{x}}_2 + \mb{N}_{\!2} \mb{J}_{\!1}^\psin \; \mb{\dot{x}}_1 \\
= \underbrace{\left[\mb{N}_{\!2}\mb{J}_{\!1}^\psin \quad \mb{J}_{\!2}^\psin\right]}_{\mb{A}^{(2)}} 
\left[\begin{matrix} \mb{\dot{x}}_1 \\ \mb{\dot{x}}_2 \end{matrix}\right]\!\!,
\label{eq:strictPriorityx2}
\end{equation}
where $ \mb{J}_{\!1}, \mb{J}_{\!2}  $ and $ \mb{N}_{\!1}, \mb{N}_{\!2} $ are the Jacobians of each task and corresponding null space projection matrices. Figure \ref{fig:planarPriorities} shows the two different prioritization strategies: priority on orientation (Figure \ref{fig:planarPrioritiesOrient}) and priority on position (Figure \ref{fig:planarPrioritiesPos}). Each demonstration is a snapshot of the robot, after satisfying the task space constraints as best as possible. In both scenarios, when the two references can be achieved simultaneously (orange and yellow references), the robot successfully fulfills the two tasks. When the two references are incompatible, the robot prioritizes one over the other, according to the demonstrated hierarchy (blue and purple references).

\begin{figure}
	\begin{subfigure}{0.47\columnwidth}
		\centering
		\includegraphics[width=1.0\columnwidth]{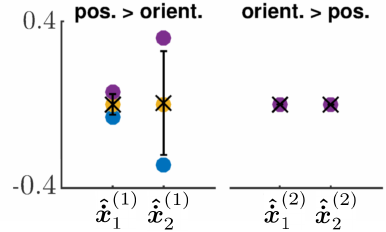}
		\caption{\centering Demonstrated \textbf{priority on orientation} (see Figure \ref{fig:planarPrioritiesOrient}).}
		\label{fig:hierarchyWeightsOrient}
	\end{subfigure}%
	\hspace{4mm}
	\begin{subfigure}{0.47\columnwidth}
		\centering
		\vspace{0.4mm}	% some alignment because images aren't exactly the same size
		\includegraphics[width=0.835\columnwidth]{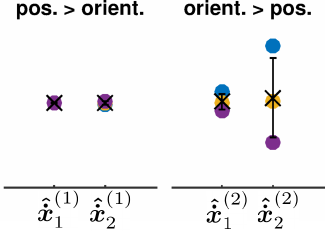}
		\caption{\centering Demonstrated \textbf{priority on position} (see Figure \ref{fig:planarPrioritiesPos}).}
		\label{fig:hierarchyWeightsPos}
	\end{subfigure}
	\vspace{0.3cm}
	\caption{Operational space velocities generated by each candidate hierarchy for the configurations of the robot in Figure \ref{fig:planarPriorities}. Each colored point corresponds to a demonstration with the same color in Figure \ref{fig:planarPriorities}. The `$ \times $' represents the mean of the datapoints, while the bars depict one standard deviation. }%The `$>$' sign in the plot titles indicates priority, i.e. the task on the left of the sign has priority over the one on the right.}
	\vspace{-0.2cm}
	\label{fig:hierarchyWeights}
\end{figure}

We propose to treat the linear operators $ \{\mb{A}^{(1)}, \mb{A}^{(2)} \}$, defined in \eqref{eq:strictPriorityx1}, \eqref{eq:strictPriorityx2} as two \textit{candidate hierarchies}, which prioritize the tasks differently. The aim is to learn which hierarchy was demonstrated, given observations of the desired operational space velocities $ \mb{\dot{x}}_1, \mb{\dot{x}}_2 $. These are equivalent to the task errors, up to a constant gain. One way to do this is through the analysis of the task space motion $ \mb{\hat{\dot{x}}}^{(j)} = \mb{J} \mb{\hat{\dot{q}}}^{(j)}$ that would result from the application of each candidate hierarchy $ j\in\{1,2\} $, i.e.,
\begin{align}
\left[\begin{matrix} \mb{\hat{\dot{x}}}^{(1)} _1 \\ \mb{\hat{\dot{x}}}^{(1)} _2 \end{matrix}\right]
& = \left[\begin{matrix} \mb{J}_{\!1} \\ \mb{J}_{\!2} \end{matrix}\right] \mb{\hat{\dot{\mb{q}}}}^{(1)} = \left[\begin{matrix} \mb{J}_{\!1} \\ \mb{J}_{\!2} \end{matrix}\right]
\left[\mb{J}_{\!1}^\psin \quad \mb{N}_{\!1}\mb{J}_{\!2}^\psin \right] 
\left[\begin{matrix} \mb{\dot{x}}_1 \\ \mb{\dot{x}}_2 \end{matrix}\right],%\nonumber
\label{eq:Task1HierarchyOp}
\end{align}
for hierarchy $ \mb{A}^{(1)} $, and
\begin{align}
\left[\begin{matrix} \mb{\hat{\dot{x}}}^{(2)}_1 \\ \mb{\hat{\dot{x}}}^{(2)}_2 \end{matrix}\right] 
& = \left[\begin{matrix} \mb{J}_{\!1} \\ \mb{J}_{\!2} \end{matrix}\right] \mb{\hat{\dot{\mb{q}}}}^{(2)} = \left[\begin{matrix} \mb{J}_1 \\ \mb{J}_2 \end{matrix}\right] 
\left[\mb{N}_2\mb{J}_1^\psin \quad \mb{J}_2^\psin \right]
\left[\begin{matrix} \mb{\dot{x}}_1 \\ \mb{\dot{x}}_2 \end{matrix}\right], %\nonumber\\
\label{eq:Task2HierarchyOp}
\end{align}
for $ \mb{A}^{(2)} $. Here, $ \mb{\dot{x}}_1, \mb{\dot{x}}_2 $ are computed from the demonstrations, given the references and the end-effector position/orientation, and they have the same value in both \eqref{eq:Task1HierarchyOp} and \eqref{eq:Task2HierarchyOp}. Figure \ref{fig:hierarchyWeights} shows the result of applying \eqref{eq:Task1HierarchyOp}, \eqref{eq:Task2HierarchyOp} to the examples of Figure \ref{fig:planarPriorities} (with $ \mb{\dot{x}}_1,\mb{\dot{x}}_2 $ computed from each task error in the depicted configuration). Notice how the datapoints exhibit low variability for the hierarchy that was demonstrated in each case (Figures \ref{fig:hierarchyWeightsOrient} and \ref{fig:hierarchyWeightsPos}). The subspaces $ j\in\{1,2\} $ associated with each candidate hierarchy $ \mb{A}^{(1)} $ and $ \mb{A}^{(2)} $, as computed from \eqref{eq:Task1HierarchyOp} and \eqref{eq:Task2HierarchyOp}, can thus be seen as containing information about the priority constraints that were demonstrated and need to be fulfilled by the robot. We thus propose to exploit the observed variability to assign importance to each candidate hierarchy. For this, we model the distribution of the data in each subspace $ j $ with a local GMM, where ${ [\mb{\xi}^\ty{I}\>\>\mb{\hat{\dot{x}}}^{(j)}] \sim \sum_{i=1}^{K}\pi_i\mathcal{N}\left(\mb{\mu}^{(j)}_i,\mb{\Sigma}^{(j)}_i\right) }$ with $ \mb{\xi}^\ty{I} $ an input. Model estimation is done using EM, as discussed in Section \ref{sec:modelEst}.
The covariance matrices $ \mb{\Sigma}^{(j)}_i $ encode the demonstrated variability in the subspace of each candidate hierarchy and % , similarly to previously discussed TP-GMM applications (Sections \ref{sec:LearningTPMovements} and \ref{sec:combiningConstraints}),
they are here exploited %(similarly to Sections \ref{sec:LearningTPMovements} and \ref{sec:combiningConstraints})
to identify priorities. In Section \ref{subsec:softWeighting} we show how $ \mb{\Sigma}^{(j)}_i $ are transformed into matrix weights for reproducing demonstrated priority behaviors in new situations.

We can easily generalize the principle used in \eqref{eq:Task1HierarchyOp} and \eqref{eq:Task2HierarchyOp} to $N_T$ arbitrary tasks and $j=1,\ldots,P$ candidate hierarchies, represented by $\mb{A}^{(j)}$. Indeed, such generalization takes the form
%
%\begin{align}
%\left[\begin{matrix} \mb{\hat{\dot{x}}}_1 \\ \vdots \\ \mb{\hat{\dot{x}}}_{N_T} \end{matrix}\right] 
%&= \left[\begin{matrix} \mb{J}_1 \\ \vdots \\ \mb{J}_{N_T} \end{matrix}\right]
%\mathcal{H}_j 
%\left[\begin{matrix} \mb{\dot{x}}_1 \\ \vdots \\\mb{\dot{x}}_{N_T} \end{matrix}\right],
%\label{eq:GeneralizedProjection}
%\end{align}
%
%where $\mathcal{H}_j$ is the candidate task hierarchy matrix $j$, built from arranging the Jacobian pseudo-inverses and null space projection matrices of the considered tasks. For the previous example, $\mathcal{H}_1 = \mb{A}^{(1)} = \left[\mb{J}_1^\psin \quad \mb{N}_1\mb{J}_2^\psin \right]$ and $\mathcal{H}_2 = \mb{A}^{(2)} = \left[\mb{N}_2\mb{J}_1^\psin \quad \mb{J}_2^\psin \right]$, but this is not always the case. Indeed, $\mathcal{H}_1 \neq \mb{A}^{(1)}$ and $\mathcal{H}_2 \neq \mb{A}^{(2)}$ if we consider the problem of tracking an object with the different priorities described by \eqref{eq:TaskParametersWithObjectTracking}. 
%
\begin{align}
\left[\begin{matrix} \mb{\hat{\dot{x}}}^{(j)} _1 \\ \vdots \\ \mb{\hat{\dot{x}}}^{(j)} _{N_T} \end{matrix}\right]
&= \left[\begin{matrix} \mb{J}_{\!1} \\ \vdots \\ \mb{J}_{\!N_T} \end{matrix}\right]
\mb{A}^{(j)} 
\left[\begin{matrix} \mb{\dot{x}}_1 \\ \vdots \\\mb{\dot{x}}_{N_T} \end{matrix}\right],
\label{eq:GeneralizedProjection}
\end{align}
%
%If we make $\mb{J}_t=\left[\mb{J}^\trsp_{t,1} \> \hdots \> \mb{J}^\trsp_{t,N_T} \right]^\trsp$, a concatenation of the Jacobian matrices of all considered tasks, and consider the datapoint $\mb{\xi}_t = \left[\mb{\dot{x}}^\trsp_{t,1} \> \hdots \> \mb{\dot{x}}^\trsp_{t,N_T} \right]^\trsp$, the mapping of the data to a frame $j$, that represents a candidate hierarchy, is obtained according to
%\begin{equation}
%\mb{X}^{(j)}_t = \mb{J}_t \mb{A}^{(j)}_t \mb{\xi}_t
%\label{eq:GeneralizedProjectionFrames}\>.
%\end{equation}
%
from which follows that local datasets are computed with ${\mb{X}^{(j)}_t = \mb{J}_t \mb{A}^{(j)}_t \mb{\xi}_t} $, where  ${\mb{J}_t=\left[\mb{J}^\trsp_{t,1} \> \hdots \> \mb{J}^\trsp_{t,N_T} \right]^\trsp}$ and ${\mb{\xi}_t = \left[\mb{\dot{x}}^\trsp_{t,1} \> \hdots \> \mb{\dot{x}}^\trsp_{t,N_T} \right]^\trsp}$. 
Note that, in this approach, we assume that the kinematic model of the robot is known, i.e., that the Jacobian of each task is available during the demonstrations, and that the set-point of each task is also given.

\subsection{Movement synthesis: fusion of strict hierarchies}
\label{subsec:softWeighting}

The most common approaches for prioritizing tasks are either based on strict hierarchies \cite{Wrede13,Saveriano15,Towell10,Hak12b}, or soft weighting of tasks \cite{Moro13,Dehio15,Modugno16,Lober15}. We propose a richer alternative based on a \textit{fusion of strict hierarchies}, gathering the best of the two approaches. For this, we exploit the learned variability in the subspace of each hierarchy, given by full covariance matrices $ \mb{\Sigma}^{(j)}_i, \forall i=1,\ldots,K , \forall j = 1,\ldots,P $ to propose a controller (dropping time subscripts $ t $ for ease of notation)
\begin{equation}
\mb{\hat{\dot{q}}} = \arg\underset{\mb{\dot{q}}}{\min} \sum_{j=1}^{P}\left(\mb{\dot{q}} - \mb{\dot{q}}^{(j)}\right)^{\!\!\trsp} \!\mb{\Gamma}^{(j)}\! \left(\mb{\dot{q}} - \mb{\dot{q}}^{(j)}\right), 
\label{eq:prioritizationOptimization}
\end{equation}
which combines candidate joint space velocities $ \mb{\dot{q}}^{(j)} $ with precision matrices $ \mb{\Gamma}^{(j)} $.
We solve the problem \eqref{eq:prioritizationOptimization} in two steps: estimating $\mb{\Gamma}^{(j)} $ and computing  $ \mb{\hat{\dot{q}}} $. For $\mb{\Gamma}^{(j)} $, GMR is first used to compute the distribution $ \mb{\hat{\dot{x}}}^{(j)}\!|\mb{\xi}^\ty{I}_t \sim \mathcal{N}\Big(\mb{\mu}^{(j)}, \mb{\Sigma}^{(j)}\Big) $ for a new input at each time step $ t $. The covariance matrix $ \mb{\Sigma}^{(j)} $, which models the importance of hierarchy $ j $ for the considered input, is then transformed with task parameters $ \mb{A}^{(j)} $ to compute a precision matrix
\begin{equation}
	\mb{\Gamma}^{(j)} = \Big(\!\mb{A}^{(j)} \mb{\Sigma}^{(j)} \mb{A}^{(j)^\trsp}\!\Big)^{\!-1}.
	\label{eq:hierarchyWeight}
\end{equation}
Note that $ \mb{\Sigma}^{(j)} $ is a squared matrix with the same number of rows and columns as the dimension of $ \mb{\dot{x}}^{(j)} $. Hence $ \mb{A}^{(j)} $ maps the covariance matrices onto joint space. $ \mb{\Gamma}^{(j)} $ are thus precision matrices that reflect the importance of each hierarchy for a given input (e.g., small values of $ \mb{\Sigma}^{(j)} $ will result in high values of $\mb{\Gamma}^{(j)} $). Then, we compute a candidate joint space velocity for each hierarchy $ j = 1,\ldots,P$ as
\begin{equation}
	\mb{\dot{q}}^{(j)} = \mb{A}^{(j)}\mb{\dot{x}}^{(j)},
	\label{eq:candidateQ}
\end{equation}
where $ \mb{\dot{x}}^{(j)}$ are the desired operational space velocities for the tasks in hierarchy $ j $.
%
%\begin{equation}
%\left\{\begin{split}
%			&\mb{\dot{q}}^{(j)}_i = \mb{A}^{(j)}\mb{\dot{x}}^{(j)}_i\\
%			&\mb{\Gamma}^{(j)}_i = \Big(\!\mb{A}^{(j)^\trsp} \mb{\Sigma}^{(j)}_i \mb{A}^{(j)}\!\Big)^{\!-1},
%		\end{split}\right.
%\end{equation}
%
Finally, it can be shown that the solution of \eqref{eq:prioritizationOptimization} corresponds to a product of $ P $ Gaussians with mean $ \mb{\dot{q}}^{(j)} $ and precision matrix $ \mb{\Gamma}^{(j)} \!\!$, i.e.,
\begin{equation}
\mb{\hat{\dot{q}}} = \Big( \sum\limits_{j=1}^P \mb{\Gamma}^{(j)} \Big)^{\!\!-1} \sum\limits_{j=1}^P \mb{\Gamma}^{(j)} \!\mb{\dot{q}}^{(j)},
\label{eq:controlQ}
\end{equation} 
similarly as in \eqref{eq:GaussProduct}. The solution $ \mb{\hat{\dot{q}}} $ is a reference joint velocity, which can be used to control the robot through a velocity or a position controller (in which case it should be integrated numerically).

Note that the complete procedure, from estimating local models based on \eqref{eq:GeneralizedProjection} to obtaining a solution \eqref{eq:controlQ}, is a modified version of TP-GMM, where the solutions for each candidate hierarchy $ \mb{\dot{q}}^{(j)} $ replace the predicted centers $  \mb{\hat{\mu}}{}^{(j)}_{t,i} $ in the original formulation \eqref{eq:GaussProduct} and $ \mb{b}^{(j)}=\mb{0} $. 
The complete approach described in Sections \ref{subsec:extractingHierarchies} and \ref{subsec:softWeighting} is summarized in Algorithm \ref{alg:learningPriorities}.

%This can be used co-jointly with a standard definition of task parameters as object-related movement. For two objects with positions $\mb{x}^\ty{O}_1, \mb{x}^\ty{O}_2$, this is achieved by defining
%%
%\begin{align}
%\mb{A}^{(1)} = \left[\mb{J}_1^\psin \quad \mb{N}_1\mb{J}_2^\psin \right] 
%,&\>\>
%\mb{b}_1 = \mb{A}^{(1)} \left[\begin{matrix} \mb{\hat{x}}_1 - \mb{x}_1 \\ \mb{\hat{x}}_2 - \mb{x}_2 \end{matrix}\right]
%,\nonumber\\
%\mb{A}^{(2)} = \left[\mb{N}_2\mb{J}_1^\psin \quad \mb{J}_2^\psin \right]
%,&\>\>
%\mb{b}_2 = \mb{A}^{(2)} \left[\begin{matrix} \mb{\hat{x}}_1 - \mb{x}_1 \\ \mb{\hat{x}}_2 - \mb{x}_2 \end{matrix}\right].
%\label{eq:TaskParametersWithObjectTracking}
%\end{align}
%%

\begin{algorithm}[bt]
	\small
	\captionsetup{font=small}
	\caption{Learning task prioritization hierarchies from demonstrations}
	\begin{algorithmic}[1]
		\Statex \textit{\textbf{Initialization}}
		\State Select a set of potentially relevant candidate hierarchies $ \mb{A}^{(j)} $ 
		\State Collect demonstrations
	\end{algorithmic}
	\begin{algorithmic}[1]
		\Statex \textit{\textbf{Model training}}
		\State Compute $ \mb{X}^{(j)} $ for each candidate hierarchy $ \mb{A}^{(j)} $ with \eqref{eq:GeneralizedProjection}
		\State Estimate model $\big\{\pi_i,\{\mb{\mu}^{(j)}_i,\mb{\Sigma}^{(j)}_i\}_{j=1}^P\big\}_{i=1}^K$ using EM
	\end{algorithmic}	
	\begin{algorithmic}[1]
		\Statex \textit{\textbf{Movement synthesis}}
		\For{$t=1,\dots,N$}
		\State Obtain input $ \mb{\xi}^\ty{I}_t $ at time step $ t $
			\For{$j=1,\dots,P$}
			\State Update task parameters $ \mb{A}^{(j)} $ with Jacobians and
			\Statex \hspace{0.95cm} null space matrices at time step $ t $
			\State Compute $ \mb{\Gamma}^{(j)} $ using GMR and  \eqref{eq:hierarchyWeight}
			\State Update desired task space velocities  $ \mb{\dot{x}}^{(j)} $ for the 
			\Statex \hspace{0.95cm} tasks in hierarchy $ j $
			\State Compute candidate solution $ \mb{\dot{q}}^{(j)}$ using \eqref{eq:candidateQ}
			\EndFor
		\State Compute velocity control command $ \mb{\hat{\dot{q}}} $ using \eqref{eq:controlQ}
		\EndFor
	\end{algorithmic}
	\label{alg:learningPriorities}
\end{algorithm}

\subsection{Experiment: Fusion of hierarchies and behavior during transitions}
An important aspect to consider when controlling priorities is the behavior of the robot when priorities change. Before testing the learning capabilities of our approach, we study the effect of varying the weights of each candidate hierarchy, when employing the controller proposed in Section \ref{subsec:softWeighting}.
\begin{figure}
	\centering	
	\includegraphics[width=\columnwidth]{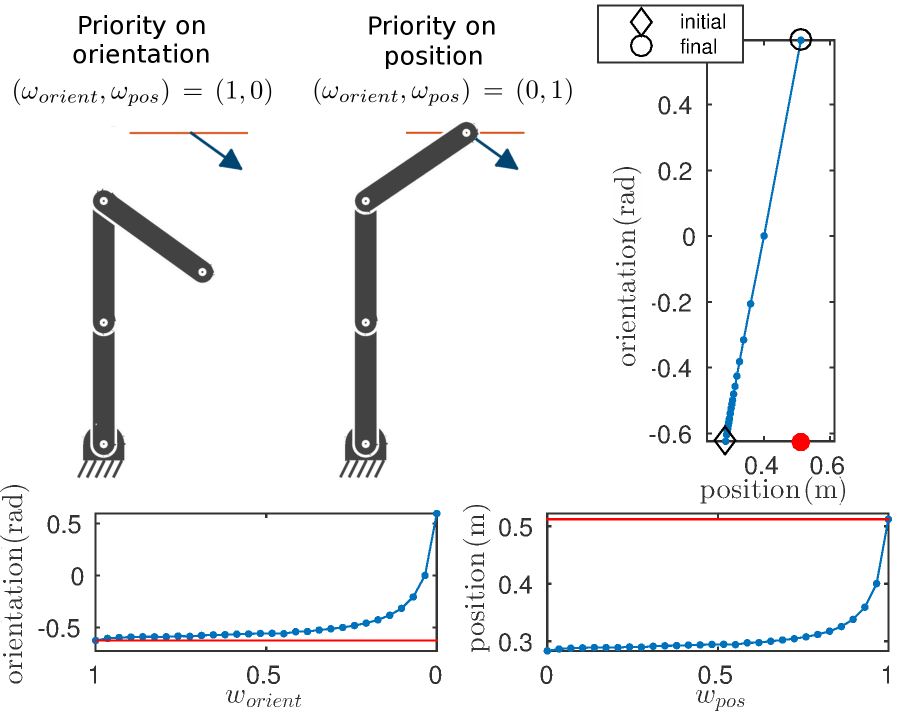}
	\caption{End-effector behavior for varying hierarchy weights. \textbf{Top-left:} Planar robot fulfilling the two tasks separately. \textbf{Top-right:} End-effector state in the position-orientation space for the considered weights (blue line). \textbf{Bottom:} End-effector orientation and vertical position as weights vary. Note the inverted scales on the horizontal axes (the experiment starts with $(1,0)$ and ends with $(0,1)$). Red lines and point indicate the references.}
	\label{fig:VariableWeights}
	\vspace{-0.0cm}
\end{figure}
\begin{figure}
	\centering
	\includegraphics[width=\columnwidth]{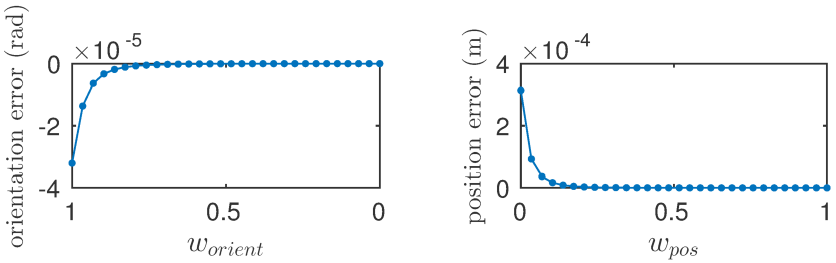}
	\caption{Performance of the robot in tracking position and orientation references with varying weights when both tasks can be achieved.
	}
	\label{fig:PlanarRobotPerformanceFeasible}
	\vspace{-0.25cm}
\end{figure}
For simplicity of analysis, we consider the case of a planar robot with two tasks. The tasks are to track a vertical position reference with the end-effector and to point downwards. The references are chosen such that the two tasks can be fulfilled individually but not simultaneously (Figure \ref{fig:VariableWeights}, \emph{top-left}). Two strict hierarchies are considered, corresponding to an accurate tracking of either position or orientation as first priority, and orientation or position as secondary. We define two scalar weights $w_{\text{pos}}$ and $w_{\text{orient}}$, which represent the importance of each hierarchy, with the subscript denoting the highest priority task. With these weights, we manually define precision matrices $\mb{\Gamma}_{\text{pos}} \!=\! w_{\text{pos}}\mb{I}$ and $\mb{\Gamma}_{\text{orient}} \!=\! w_{\text{orient}}\mb{I}$, where $\mb{I}$ is a 2-dimensional identity matrix. Figure \ref{fig:VariableWeights} shows the results for the case in which the robot transits from priority on orientation to priority on position, through a continuous variation of the weights. We observe that, as different combinations of weights are applied by the controller, a smooth transition occurs between the two tasks\footnote{Note that the transition is non-linear, unlike the variation of the weights $w_{\text{pos}}$, $w_{\text{orient}}$. This is because the Jacobians of each task, involved in \eqref{eq:hierarchyWeight}, change non-linearly with the robot configuration $ \mb{q} $.}. These results show that the proposed controller can handle situations where priorities change, which can occur when demonstrated hierarchies vary over time or according to different contexts. 

Another case that is worth investigating is the behavior of the system when both tasks are achievable. In this situation, one would expect that any combination of the weights would result in a successful completion of both tasks. In order to test how our approach handles this scenario, we decreased the vertical position reference so that both tasks can be fulfilled simultaneously.
We set the robot to an initial configuration ${\mb{q}_{\text{init}} = [\frac{\pi}{2}\!+\!0.5, \>\> -1, \>\> -\frac{\pi}{2}\!+\!0.5]}$. For every new weight pair, its starting configuration is that of the previous pair. Figure \ref{fig:PlanarRobotPerformanceFeasible} shows the tracking performance. We can see that both position and orientation were tracked with negligible errors. This shows that the approach converges to a proper solution and maintains it when the weights are modified. In other words, once the robot converged to a configuration that fulfills both tasks with the initial weights $(w_{\text{orient}},w_{\text{pos}}) = (1,0)$, further variations of the weights did not affect the solution, as one would expect. 

\subsection{Experiment: Learning priorities with the COMAN robot}
\label{sec:expCOMAN}

\begin{figure}
	\centering
	\includegraphics[width=\columnwidth]{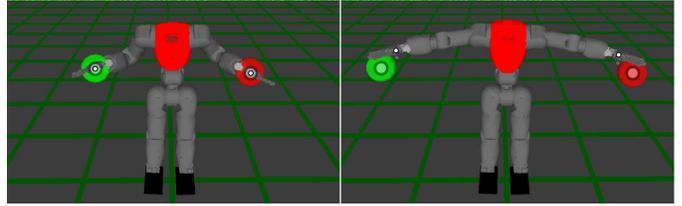}
	\caption{Illustration of the waist joint prioritization problem. \textbf{Left:} Both targets are within the reachable workspace of both arms, so the robot can successfully reach the centers of the spheres with the palm of its hands. \textbf{Right:} References become simultaneously unreachable, so the robot cannot reach any of the two. The small circles depict the tool center point of each arm, while the large circles represent the centers of the references.}
	\label{fig:COMAN_PrioritiesCloseFar}
	\vspace{0.1cm}
\end{figure}
\begin{figure}%[h]
	\begin{subfigure}{0.49\columnwidth}
		\centering
		\includegraphics[width=1.0\columnwidth]{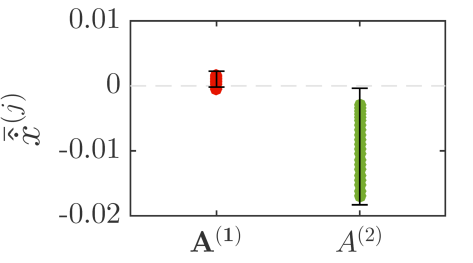}
		\caption{Demonstrated \textbf{left arm priority}.}
		\label{fig:PriorityDataSetsLeftArm}
	\end{subfigure}
	\begin{subfigure}{0.49\columnwidth}
		\centering
		\includegraphics[width=0.99\columnwidth]{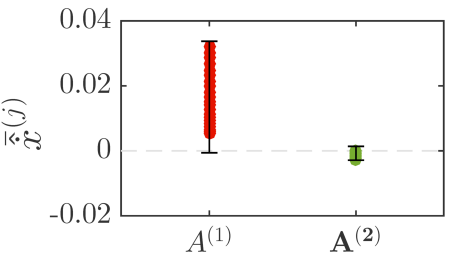}
		\caption{Demonstrated \textbf{right arm priority}.}
		\label{fig:PriorityDataSetsRightArm}
	\end{subfigure}
	\vspace{0.3cm}
	\caption{Task space velocities generated by each candidate hierarchy given the demonstration data. Each entry $ \mb{A}^{(j)} $ contains the means of several observations $ \mb{\hat{\dot{x}}}^{(j)}$, computed across the dimensions of the tasks. The error bars correspond to two standard deviations of the data. Note the low variability in the datapoints corresponding to the demonstrated hierarchies.}
	\label{fig:PriorityDataSets}
	\vspace{-0.1cm}
\end{figure}

We now test the hierarchy identification and synthesis capabilities of the approach with the humanoid robot COMAN. Since the kinematic chains of the humanoid arms share the waist joints, incompatibilities often occur during bimanual manipulation when the tasks of the two arms are too distant (Figure \ref{fig:COMAN_PrioritiesCloseFar}). In this experiment we use the proposed approach to teach the robot which arm has priority over the other. If correctly learned, the robot will prioritize one of the references when the two are incompatible and closely track both when reachable. We consider $P=2$ candidate hierarchies, corresponding to the two possible combinations of priorities, i.e., priority on left arm with the right arm as secondary and vice-versa. We denote the Jacobians of each arm (from the waist link) by $\mb{J}_{\!\!L}$, $\mb{J}_{\!\!R}$, with corresponding pseudoinverses $\mb{J}^\psin_{\!\!L}$, $\mb{J}^\psin_{\!\!R}$ and null space projection matrices  $\mb{N}_{\!\!L}$, $\mb{N}_{\!\!R}$, yielding the task parameters
\begin{align}
\mb{A}^{(1)} = \left[\mb{J}^\psin_{\!\!L} \>\>\> \mb{N}_{\!\!L}\mb{J}^\psin_{\!\!R} \right],\qquad
\mb{A}^{(2)} = \left[\mb{N}_{\!\!R}\mb{J}^\psin_{\!\!L} \>\>\> \mb{J}^\psin_{\!\!R} \right],
\label{eq:TPsWithObjectTrackingBimanual}
\end{align}
and the desired task space velocities for each hierarchy
\begin{equation}
	\mb{\dot{x}}^{(1)} = \mb{\dot{x}}^{(2)} = \left[\begin{matrix} \mb{\hat{x}}_L - \mb{x}_L \\ \mb{\hat{x}}_R - \mb{x}_R\end{matrix}\right],
\end{equation}
where $\mb{\hat{x}}_L$ and $\mb{\hat{x}}_R$ denote left and right reference positions.
For model training, we consider datapoints of the form ${\mb{\xi}_t = \left[\begin{matrix} \mb{\dot{x}}^\trsp_{L,t} \>\> \mb{\dot{x}}^\trsp_{R,t}\end{matrix}\right]^\trsp \in \mathbb{R}^6}$, where $\mb{\dot{x}}_{L,t} = \mb{\hat{x}}_{L,t}-\mb{x}_{L,t} $ and $ \mb{\dot{x}}_{R,t} = \mb{\hat{x}}_{R,t}-\mb{x}_{R,t}$ are the desired left and right end-effector velocities during demonstrations. Equation \eqref{eq:GeneralizedProjection} is used to compute the local datasets $ \mb{X}^{(j)} $. We choose $K=1$, i.e., we want to have a constant prioritization structure throughout the task. This is to show that we can already encode complex behaviors with a single Gaussian. More complex structures can be easily considered by adding more components (e.g., to teach input dependent priorities).
\begin{figure}%[h]
	\centering
	\includegraphics[width=0.49\columnwidth]{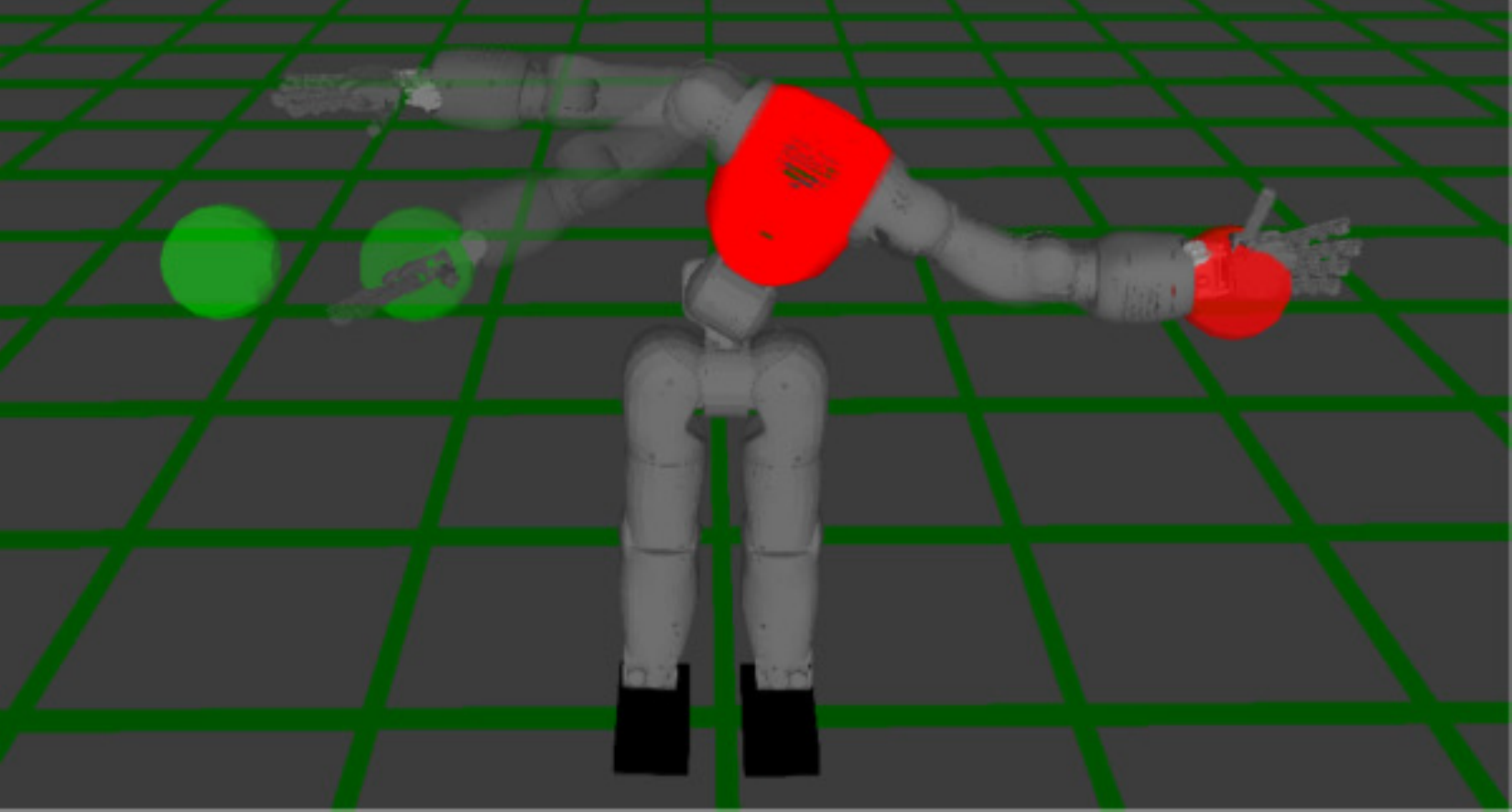}
	\includegraphics[width=0.49\columnwidth]{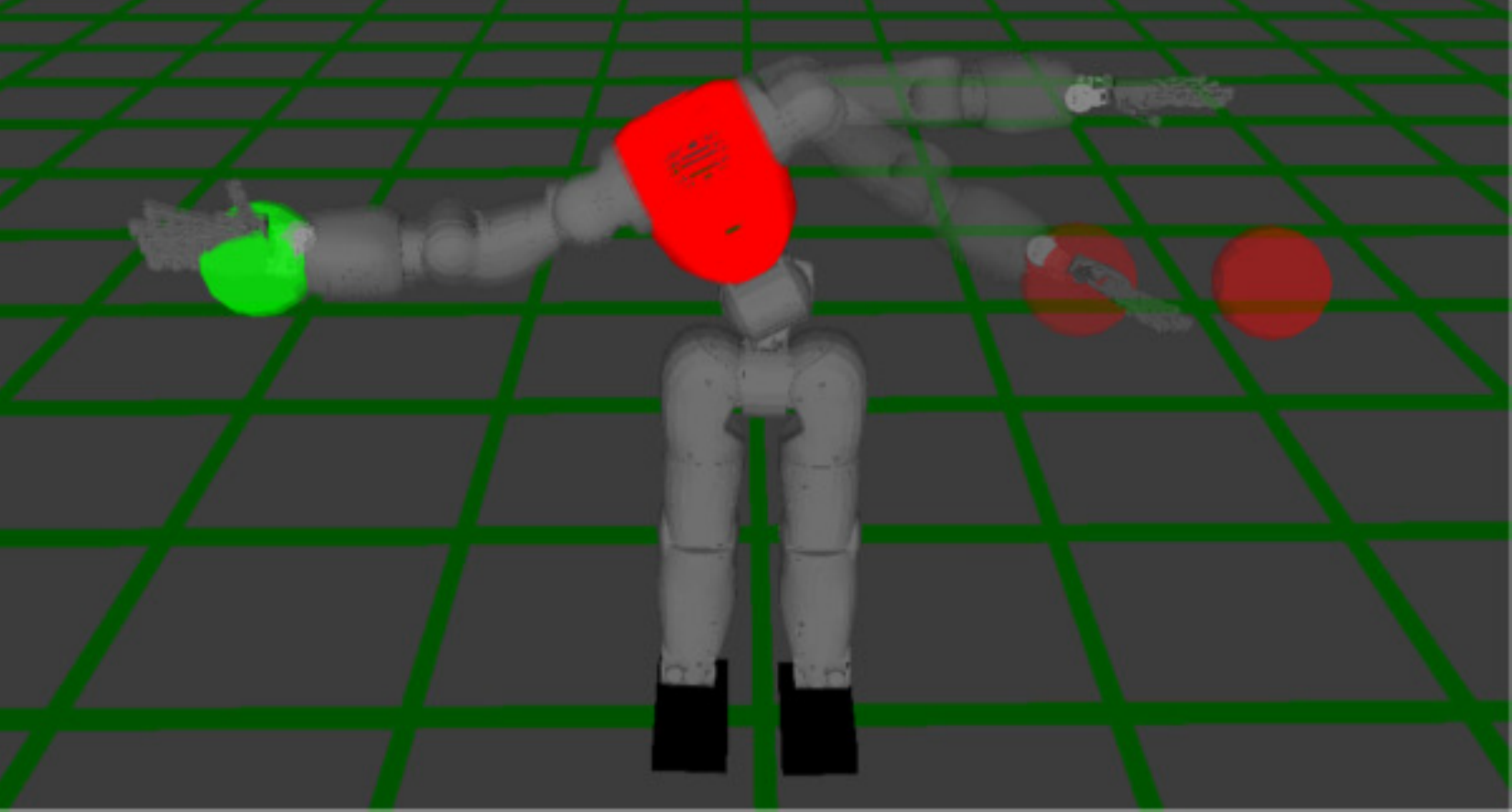}
	\caption{Movement synthesis using the learned left and right arm priority models. \textbf{Left:} The left reference is always tracked, while the right reference is only tracked when reachable. \textbf{Right:} The right reference is prioritized.}
	\label{fig:COMANPrioritySim}
	\vspace{0.2cm}
\end{figure}
\begin{figure}%[h]
	\centering
	\includegraphics[width=\columnwidth]{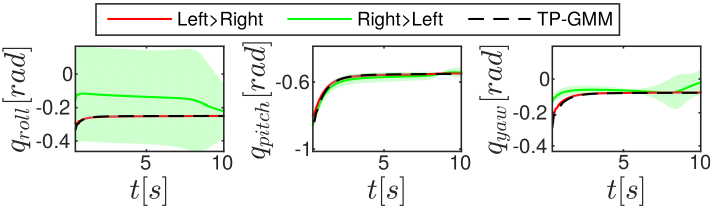}
	\caption{Waist joint angles given by each candidate hierarchy. The envelope around the lines represents one standard deviation of the variance encoded in each hierarchy. TP-GMM extracts the solution with the lowest variability. }%The `$>$' sign in the plot legends indicates priority, i.e. the task on the left of the sign has priority over the one on the right.}
	\label{fig:COMAN_WaistJointsLeft}
	\vspace{-0.0cm}
\end{figure}

\begin{figure*}
	\centering
	\includegraphics[width=1.0\textwidth]{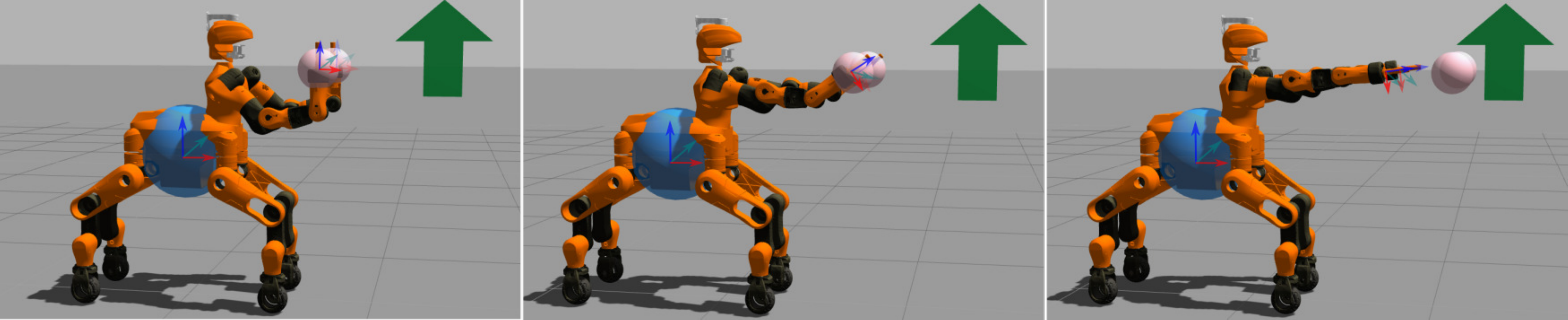}
	\caption{The Centauro robot tracks different references, prioritizing (by decreasing order of importance) base position, end-effector positions and end-effector orientations. The blue and pink spheres represent the desired base and end-effectors positions, while the arrow denotes the desired end-effector pointing direction. \textbf{Left:} All tasks can be fulfilled. \textbf{Center:} The pelvis reference is moved backwards, thus the robot performs the orientation task as best as it can but not completely. \textbf{Right:} The end-effector position references are moved forward and both end-effector position and orientation are now only partially fulfilled.}
	\label{fig:CentauroPriorityDemos}
	\vspace{-0.2cm}
\end{figure*}

Demonstrations of the priority behaviors are given by implementing an inverse kinematics controller with the desired hierarchies in the simulated robot and making it track moving references\footnote{Alternatively, kinesthetic teaching or optical tracking of movements from humans could also be used by recording the tracking errors as well as the kinematic data of the demonstrator (required to compute the Jacobians).}. %During each demonstration, the priority behavior was shown, i.e., one of the arms could not fulfill its task while the other could. 
We generate demonstrations of two different types of priority and analyzed the resulting datasets. Figure \ref{fig:PriorityDataSets} shows the mean of $ \mb{\hat{\dot{x}}}^{(j)} $, generated by each candidate hierarchy, computed from \eqref{eq:GeneralizedProjection} with task parameters \eqref{eq:TPsWithObjectTrackingBimanual}. As one would expect from Section \ref{subsec:extractingHierarchies}, we observe a distribution of datapoints with low variability for the hierarchy that was demonstrated, with priority either on the left arm (Figure \ref{fig:PriorityDataSetsLeftArm}) or on the right arm (Figure \ref{fig:PriorityDataSetsRightArm}). Low variability results in high precision matrices $\mb{\Gamma}^{(j)}$, which in turn, during movement synthesis, result in higher weights to favor the corresponding hierarchy.

\begin{figure}
	\begin{subfigure}{0.49\columnwidth}
		\centering
		\includegraphics[width=\columnwidth]{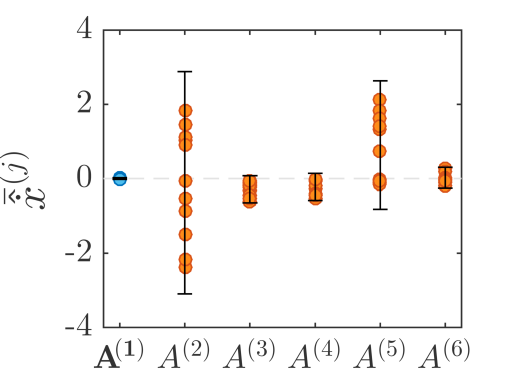}
		%		\caption{\centering \textbf{Demonstrated:} base position $ > $ end-effector positions $ > $ end-effector orientations}
		\caption{\centering \textbf{Demonstrated:} $b > p > o$.}
		\label{fig:PrioPelvisPosOrient}
	\end{subfigure}
	\begin{subfigure}{0.49\columnwidth}
		\centering
		\includegraphics[width=\columnwidth]{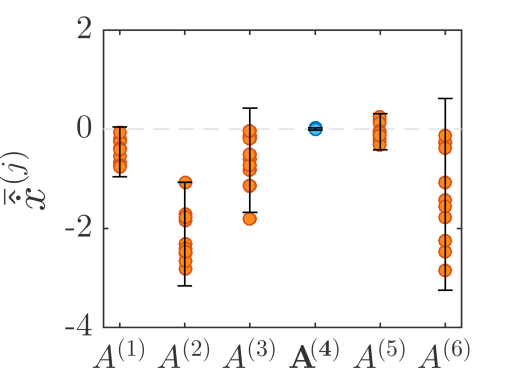}
		\caption{\centering \textbf{Demonstrated:} $p > b > o$.}
		\label{fig:PrioPosPelvisOrient}
	\end{subfigure}
	\vspace{0.3cm}
	\caption{ Datasets from Centauro priority demonstrations. Each entry $ \mb{A}^{(j)} $ contains 12 points (the means of $ \mb{\hat{\dot{x}}}^{(j)} $ for each demonstration).}% The `$>$' sign in the subcaptions indicates priority, i.e. the task on the left of the sign has priority over the one on the right.}
	\label{fig:demoDataCentauro}
	\vspace{-0.0cm}
\end{figure}
\begin{figure}
	\centering
	\vspace{0.25cm}
	\includegraphics[width=0.8\columnwidth]{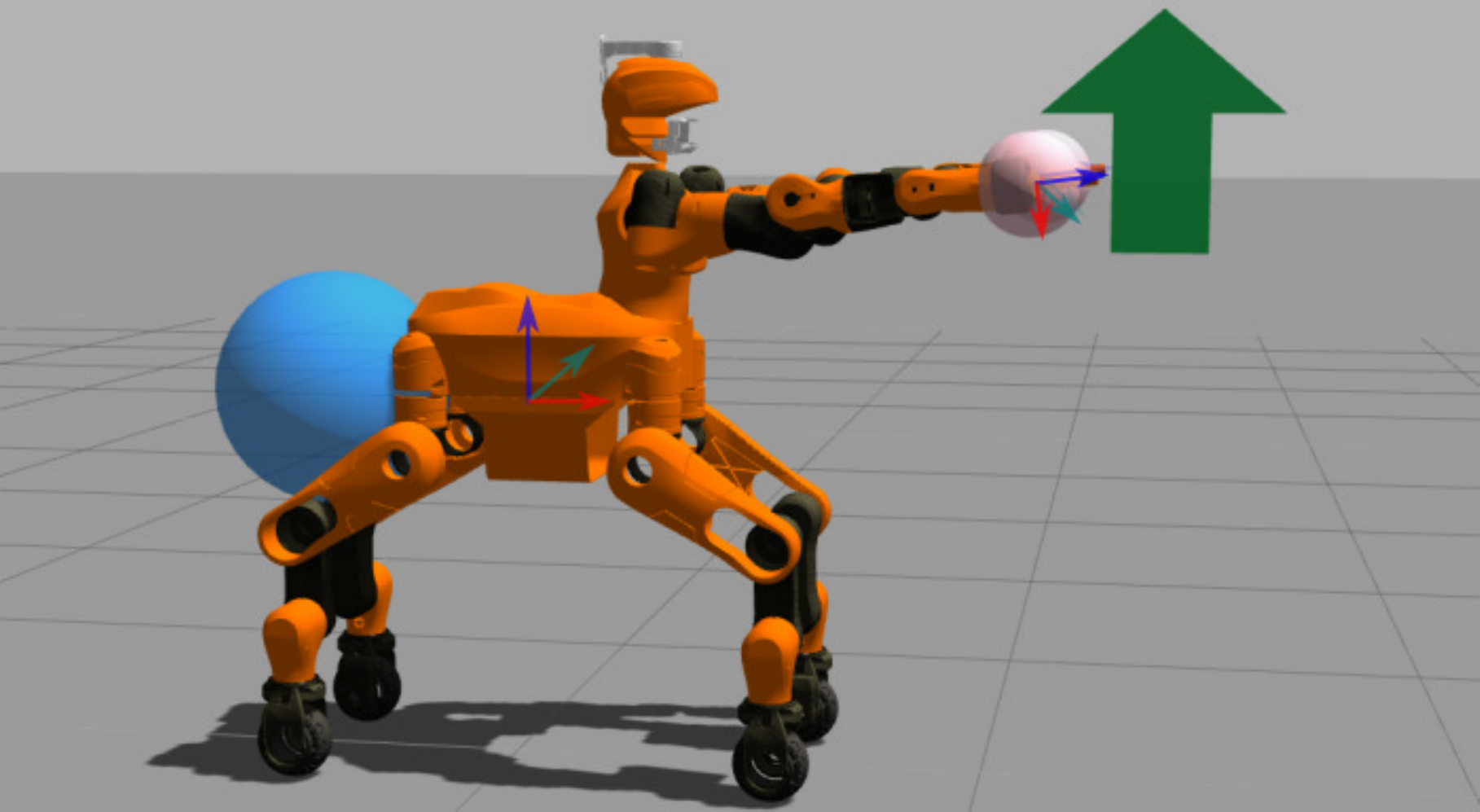}
	\caption{ In this example, the Centauro robot prioritizes end-effector positions, followed by the position of the base and, finally, the end-effector orientations.}
	\label{fig:CentauroPriorityEndEff}
	\vspace{-0.cm}
\end{figure}

We apply the learned models (left and right arm priority) to reach fixed points on the left and right sides of the robot, with the results depicted in Figure \ref{fig:COMANPrioritySim}. %Figures \ref{fig:COMANPriorityLeft} and \ref{fig:COMANPriorityRight}
As expected, in both cases, we observe that the arm which had the highest priority during demonstrations always fulfills the task, while the other only does it when the reference is reachable. Figure \ref{fig:COMANPriority} shows snapshots of the same behaviors in the real COMAN platform, where we used an optical tracking system to provide moving references to each arm. The robot closely tracked the reference on the side that had the highest priority, while doing its best to track the secondary reference. The computations in rows 2--9 of Algorithm \ref{alg:learningPriorities} (computation at each time step) took on average $ 1.15 ms $, using MATLAB in a laptop with an Intel Core i5-3230M, {$ 2.60\mathrm{GHz} \times 4$} processor. Videos of the experiment can be found at \mbox{\texttt{\footnotesize\url{http://joaosilverio.weebly.com/tro}}}.

In this framework, similarly to the experiment in Section \ref{subsec:shakingTask}, each candidate hierarchy provides one possible configuration space solution. %For a humanoid performing a bimanual skill, the solutions differ in how the waist joints are prioritized (since they are shared by the arms).
Figure \ref{fig:COMAN_WaistJointsLeft} shows the solution given by each candidate hierarchy for the waist roll, pitch and yaw joints, while tracking the left object with the highest priority. We see that the solution given by the hierarchy that encodes priority on the left arm (red line) has much lower variability than the one that prioritizes tracking with the right arm (green line). This leads to an automatic identification of this prioritization as being important, resulting in a proper reproduction of priorities.

%\subsection{Experiment: Identification and synthesis of hierarchies with more than two tasks}
\subsection{Experiment: Learning priorities with more than two tasks}
\label{sec:expCentauro}

\renewcommand{\arraystretch}{1.3}
\begin{table}
	\begin{tabular}{c|c|c|c|c|c}
			$ \mb{A}^{(1)} $ & $ \mb{A}^{(2)} $ & $ \mb{A}^{(3)} $ & $ \mb{A}^{(4)} $ & $ \mb{A}^{(5)} $ & $ \mb{A}^{(6)} $ \\
			\hline
			$ b\!>\!p\!>\!o $ & $ b \!> \!o \!> \!p $ & $ p \!> \!o > \!b $ & $  p \!>\! b \!>\! o $ & $ o \!>\! p \!>\! b $ & $ o \!>\! b \!>\! p $ \\
		\end{tabular}
		\caption{Hierarchies used in Section \ref{sec:expCentauro}.}
		\label{tab:prioritiesCentauro}
\end{table}
\renewcommand{\arraystretch}{1.0}

In this experiment we consider a loco-manipulation task with the new Centauro robot from IIT \cite{Baccelliere2017}. Centauro is a four-legged robot with wheels and a 15-DOF humanoid torso. In order to simplify the locomotion problem, %(e.g. not consider the control of the center of mass), 
we constrain the robot to move using only the wheels, becoming, in essence, a mobile platform with a humanoid torso. We consider three tasks that the robot can execute at the same time, corresponding to the control of:
\begin{compactitem}
	\item the \textit{position} of the robot's \textit{base}, assumed to be the pelvis link, with associated Jacobian and null space projection matrices $ \mb{J}_{b}, \mb{N}_{b} $,
	\item the \textit{positions} of \textit{both end-effectors}, with matrices $ \mb{J}_{p}, \mb{N}_{p} $,
	\item the \textit{orientations} of \textit{both end-effectors}, with $ \mb{J}_{o}, \!\mb{N}_{o}. $
\end{compactitem}

The possible arrangements of these three tasks yield 6 candidate hierarchies (see Table \ref{tab:prioritiesCentauro} where we used the Jacobian subscripts to denote each task). In many contexts, one may know in advance that the robot will only employ a subset of the total available hierarchies (e.g., one may assume that the pelvis position is not an important sub-task). We here consider all possible combinations with $ P=6 $, with the aim of showing the potential of the approach to resolve tasks with a complete set of candidate hierarchies. In this experiment, we control a total of 19 joints: 4 wheels and 15 upper body joints (7 per arm and 1 torso joint).

Using the Gazebo simulator, we collect 12 demonstrations of the robot employing the hierarchy $ {\mb{A}^{(1)} = [\mb{J}_{b}^\dagger \>\>\>\> \mb{N}_{b}\mb{J}_{p}^\dagger  \>\>\>\>\mb{N}_{b} \mb{N}_{p} \mb{J}_{o}^\dagger ]}$, which controls the base position with the highest priority, followed by the end-effector positions and, finally, orientations.  %Note that while these demonstrations were collected from Gazebo, it would be straightforward to collect them from the real robot, provided knowledge of the references, which can be obtained through vision feedback or optical tracking.
Each demonstration uses different combinations of the position references (base and end-effectors), while keeping the orientation reference the same, in this case the quaternion $ \mb{\hat{\epsilon}} = [0 \>\> 0 \>\> 0 \>\> 1]^\trsp $. Figure \ref{fig:CentauroPriorityDemos} shows 3 demonstration examples, for different values of the references to show the priority behaviors. The demonstration data is stored when the robot is at a configuration where each task is fulfilled as best as possible, according to the hierarchy. Each demonstration results in $ P $  vectors $ \mb{\hat{\dot{x}}}^{(j)} $, as computed from \eqref{eq:GeneralizedProjection}.

Figure \ref{fig:PrioPelvisPosOrient} shows the mean of the resulting vectors for each $ j=1,\ldots,P$, with error bars denoting two standard deviations. Notice that the demonstrated hierarchy generated the lowest variance. This confirms our proposition from Section \ref{subsec:extractingHierarchies}, validating the approach for the extraction of hierarchies with more than two tasks. The reproduction of the learned priorities, for the same and different references, is shown in the video that can be found at \mbox{\texttt{\footnotesize\url{http://joaosilverio.weebly.com/tro}}}. Also note that, in Fig.\ \ref{fig:PrioPelvisPosOrient}, we take the mean of $ \mb{\hat{\dot{x}}}^{(j)} $ only for visualization purposes. The reproduction of priority behaviors is done using \eqref{eq:controlQ}, which takes into account the variability of each dimension separately. For this 6-hierarchy experiment, the operations in rows 2--9 of Algorithm \ref{alg:learningPriorities} (computation at each time step) took on average $ 7.35ms $ per control cycle, in a computer with an Intel Core i7-6700, ${3.40\mathrm{GHz} \times 8}$ processor using a C++ (unoptimized) implementation based on the Eigen library for linear algebra computations. The increase in computational time with respect to the experiment reported in Section \ref{sec:expCOMAN} is due to a higher number of hierarchies and controlled degrees of freedom (we considered the 48-dimensional whole-body Jacobian), which increases the cost of the inversions in \eqref{eq:hierarchyWeight} and \eqref{eq:controlQ}.

In order to show that the approach can successfully extract any demonstrated hierarchy, we provide 12 new demonstrations of a different hierarchy, in this case $ {\mb{A}^{(4)} = [\mb{N}_{p}\mb{J}_{b}^\dagger \>\>\>\> \mb{J}_{p}^\dagger  \>\>\>\>\mb{N}_{p} \mb{N}_{b} \mb{J}_{o}^\dagger]}$, where the end-effector positions are prioritized, followed by the pelvis position and end-effector orientations. Figure \ref{fig:CentauroPriorityEndEff} shows an example of such hierarchy. In Fig.\ \ref{fig:PrioPosPelvisOrient}, we see that the demonstrated hierarchy is properly extracted, since the data show low variability for $ \mb{A}^{(4)} $. The accompanying video shows the reproduction of this hierarchy, highlighting how it differs from $ \mb{A}^{(1)} $.

\subsection{Experiment: Transferring task priorities between robots}

\begin{figure}%[h]
	\centering
	\includegraphics[width=0.49\columnwidth]{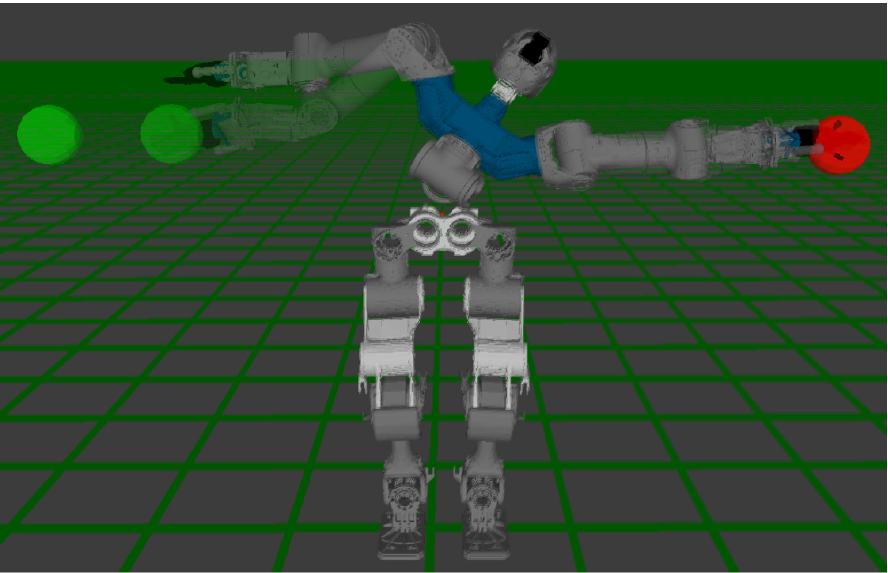}
	\includegraphics[width=0.49\columnwidth]{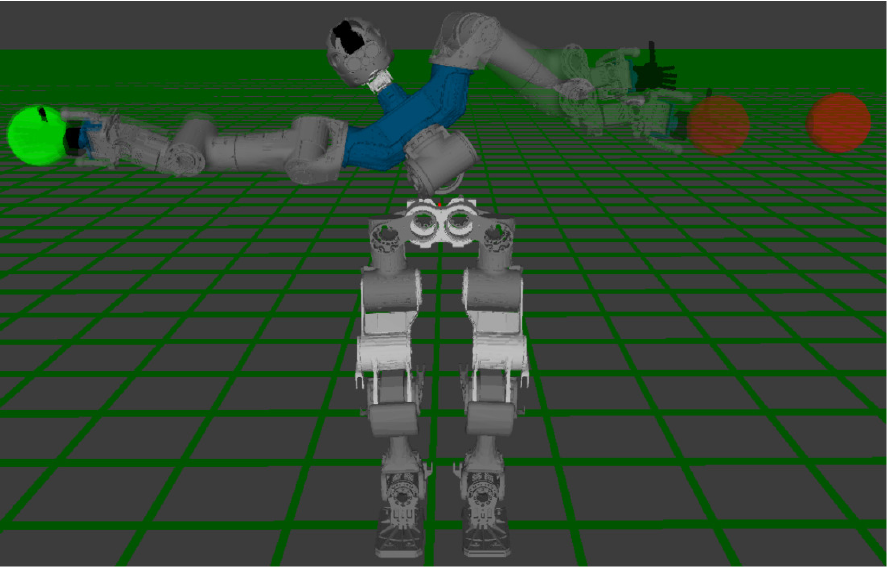}
	\caption{Movement synthesis with the WALK-MAN robot using the left and right priority models that were learned with COMAN.}
	\label{fig:WALKMANPriority}
	\vspace{-0.2cm}
\end{figure}

In this experiment we test the transfer of priority models across different robots. We use the model learned with COMAN to synthesize an equivalent prioritization behavior in WALK-MAN \cite{Tsagarakis17} that has different kinematic parameters. WALK-MAN is a humanoid larger than COMAN, whose waist joints are also shared between the arms. Figure \ref{fig:WALKMANPriority} shows that the model acquired using COMAN is successful in generalizing the learned task priorities in WALK-MAN. This stems from two facts: i) the considered tasks, and associated weights, are in operational space, whose dimension is the same for both robots and ii) the robots have the same kinematic chains structure, i.e., waist joints are shared by both arms. 
%(it would not make sense to apply this priority model, for example, in a dual-arm platform of two single arms). 
This opens up an interesting strategy to cope with the correspondence problem in imitation learning. In particular, our approach can potentially be used by the robot to visually observe a task prioritization behavior demonstrated by the user without considering a direct mapping of the kinematics.

%\section{Advantages and Limitations of the approach} %Shortcomings of the approach
\section{Discussion}
\label{sec:discussion}

\subsection{Learning operational and configuration space constraints}
Our results in Section \ref{sec:combiningConstraints} showed that, with Jacobian-based task parameters, the robot can take into account both configuration and operational space constraints, including orientation. Learning controllers for complex bimanual manipulation goes beyond operational space, and one must also care about the configuration space to achieve natural and efficient movements. The shaking experiment is an example of task that is best modeled by taking into account the configuration space constraints, but the range of possibly interesting situations is vast (e.g., communicative gestures, self-collision avoidance). %Usually, in control, one can define this kind of information \textit{a priori}, but having it learned from demonstrations eases the planning.
Although highly relevant, this problem is rarely covered in the literature of motor skill learning, especially in bimanual cases. %, but we understand that it is highly relevant. %We have therefore provided a possible way of addressing the challenging problem of learning and combining constraints in both spaces. 

One potential shortcoming of our approach is that it relies on the variability of the demonstrated movements to extract the importance of each space. In some cases, such task variations may not be straightforward to demonstrate (e.g., unexperienced demonstrators may not demonstrate sufficient variability). One possible way to cope with this issue could be to combine TP-GMM with more interactive and incremental teaching approaches, which would allow the robot to refine its task model at runtime by testing task variations in the different spaces based on exploration, interaction and teacher feedback. 

In a different direction, we plan to study how the proposed approach can be combined with Riemannian formulations for learning end-effector poses \cite{Zeestraten2017}. While here we approximate the geometry of the unit quaternion space, such formulations may allow for a more precise encoding of orientation.

Sections \ref{sec:combiningConstraints} and \ref{sec:priorityConstraints} both considered the combination of sub-spaces as a fusion problem with full precisions matrices, which has a richer structure than a combination of sub-spaces through scalar weights. In particular, they both allow a parallel organization of tasks by exploiting the sparsity of the sub-space constraints. Although better than a superposition through scalar weights, such organization still cannot guarantee that all sub-space constraints can be fulfilled when these constraints are not compatible. Currently, the robot finds a trade-off without evaluating whether this solution is satisfactory or whether sub-spaces are too much in conflict to be fused properly. It would be interesting to investigate in future work if such cases could be detected and how the robot should cope with highly conflicting constraints.

\subsection{Learning task prioritization hierarchies}
By extending the notion of candidate coordinate systems to candidate task hierarchies, we showed that TP-GMM could be used to learn priority hierarchies. 

One potential shortcoming is that the set of possible task hierarchies needs to be set beforehand. To a certain extent, one can take advantage of expert knowledge to only use a subset of all possible candidate hierarchies. However, even though one is not required to provide all possible hierarchies for the considered tasks, one should be careful not to over-define this set, because the more candidate hierarchies are available, the more demonstrations will be required to extract invariant features through statistics. A potential research direction to leverage this issue could be to study ways of learning sets of hierarchies characterizing specific domains of activity (e.g., bimanual manipulation, walking). Then, one could take advantage of the prior knowledge about the skill to be demonstrated to select the set accordingly.

Another potential drawback is the need to define the set-points that the robot may be tracking. Note that this is a common hypothesis in approaches extracting task prioritization, see e.g., \cite{Towell10, Lin15}. While this may not pose a problem in simpler scenarios (e.g., the targets do not move), the difficulty increases in highly dynamic environments. One intuitive solution to overcome this limitation could be to split the demonstration of the task in two phases (an idea exploited in \cite{Wrede13} albeit in a more simple case). In one phase, each individual sub-task could be demonstrated (e.g., pelvis trajectory or end-effector movements with respect to objects) and encoded in separate models. In the second phase, the environment could be simplified (e.g., fixing references) and the priorities demonstrated. Equations \eqref{eq:candidateQ}, \eqref{eq:controlQ} ensure a flexible solution at this level, since the control commands $ \mb{\dot{x}}^{(j)} $ of each sub-task in \eqref{eq:candidateQ} may be generated from a model that is trained independently from the priority weights used in \eqref{eq:controlQ}.

%%%%%%%%%%%%%%%%%%%%%%%%%%%%%%%%%%%%%%%%%%%%%%%%%%%%%%%%%%%%%%%%%%%%%%%%%%%%%%%%%%%%%%%%%%%%%%%%%%%%%%
% VII. Conclusions and Future Work
%%%%%%%%%%%%%%%%%%%%%%%%%%%%%%%%%%%%%%%%%%%%%%%%%%%%%%%%%%%%%%%%%%%%%%%%%%%%%%%%%%%%%%%%%%%%%%%%%%%%%%

\section{Conclusions and Future Work}
\label{sec:conclusions}
We presented a framework for human-robot bimanual skill transfer based on Task-Parameterized Gaussian Mixture Models. We introduced different parameterizations allowing us to consider a wide range of learning problems related to manipulation in humanoids, namely the combination of constraints between operational and configuration spaces, and the learning of task priority hierarchies. 
The approach was validated in three different scenarios. We first demonstrated, in a bimanual shaking task with the COMAN robot, that the approach can be used to consider constraints in operational and configuration spaces simultaneously, which permits the reproduction of motion patterns from both spaces during movement synthesis. Then, in a bimanual reaching experiment with COMAN, we showed that by providing different candidate hierarchies to describe an observed movement, the robot is able to determine from statistics which prioritization is relevant for the task. In this context, we also demonstrated that the proposed formulation could be employed to transfer prioritization behaviors between humanoids. Finally, in a loco-manipulation scenario with the Centauro robot, we showed that our framework can be used to learn the prioritization of more than two simultaneous tasks from demonstrations.

Future work will investigate how other types of constraints could be formulated for torque-controlled robots \cite{Silverio2018b}, as opposed to velocity controllers considered in this article. A growing number of robots can be controlled in torque, thus it is relevant for learning approaches to account for this fact.

We also plan to investigate if the proposed approach could be combined with other optimization-based techniques, such as CMA-ES \cite{Dehio15,Modugno16}, in order to refine the learned priorities. This would allow us to consider cost functions that are harder to relate to operational space control, such as energy efficiency. 

A final research direction concerns the application of task prioritization with TP-GMM to more complex whole-body control scenarios. One possible avenue could be that of learning whole body motion behaviors from human demonstrations, including prioritization skills involving center of mass and contact points with the environment (e.g., to learn natural ways of coping with perturbations while standing).

\ifCLASSOPTIONcaptionsoff
  \newpage
\fi

\bibliographystyle{bibtex/IEEEtran}
\bibliography{bibtex/IEEEabrv,bibtex/bibTRO}

\begin{IEEEbiography}[{\includegraphics[width=1in,height=1.25in,trim=15 0 15 0,clip,keepaspectratio]{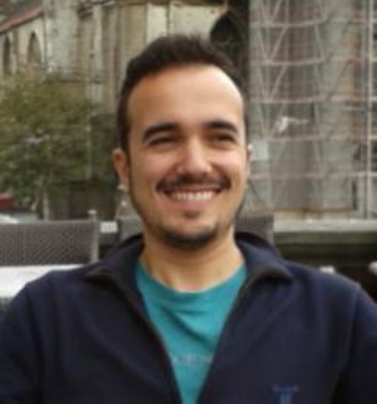}}]{Jo\~ao Silv\'erio} is a Postdoctoral Researcher at the Department of Advanced Robotics, Istituto Italiano di Tecnologia (IIT) since June 2017. He received his M.Sc in Electrical and Computer Engineering (2011) from Instituto Superior T\'ecnico (Lisbon, Portugal) and Ph.D in Robotics (2017) from the University of Genova (Italy) and IIT. Before his PhD, he also carried out research at EPFL's Biorobotics Laboratory in 2010 and Learning Systems and Algorithms Laboratory in 2013. He is interested in robot programming by demonstration, in particular algorithms for learning complex bimanual skills. Webpage: \texttt{\footnotesize http://joaosilverio.eu}
\end{IEEEbiography}

\begin{IEEEbiography}[{\includegraphics[width=1in,height=1.25in,clip,keepaspectratio]{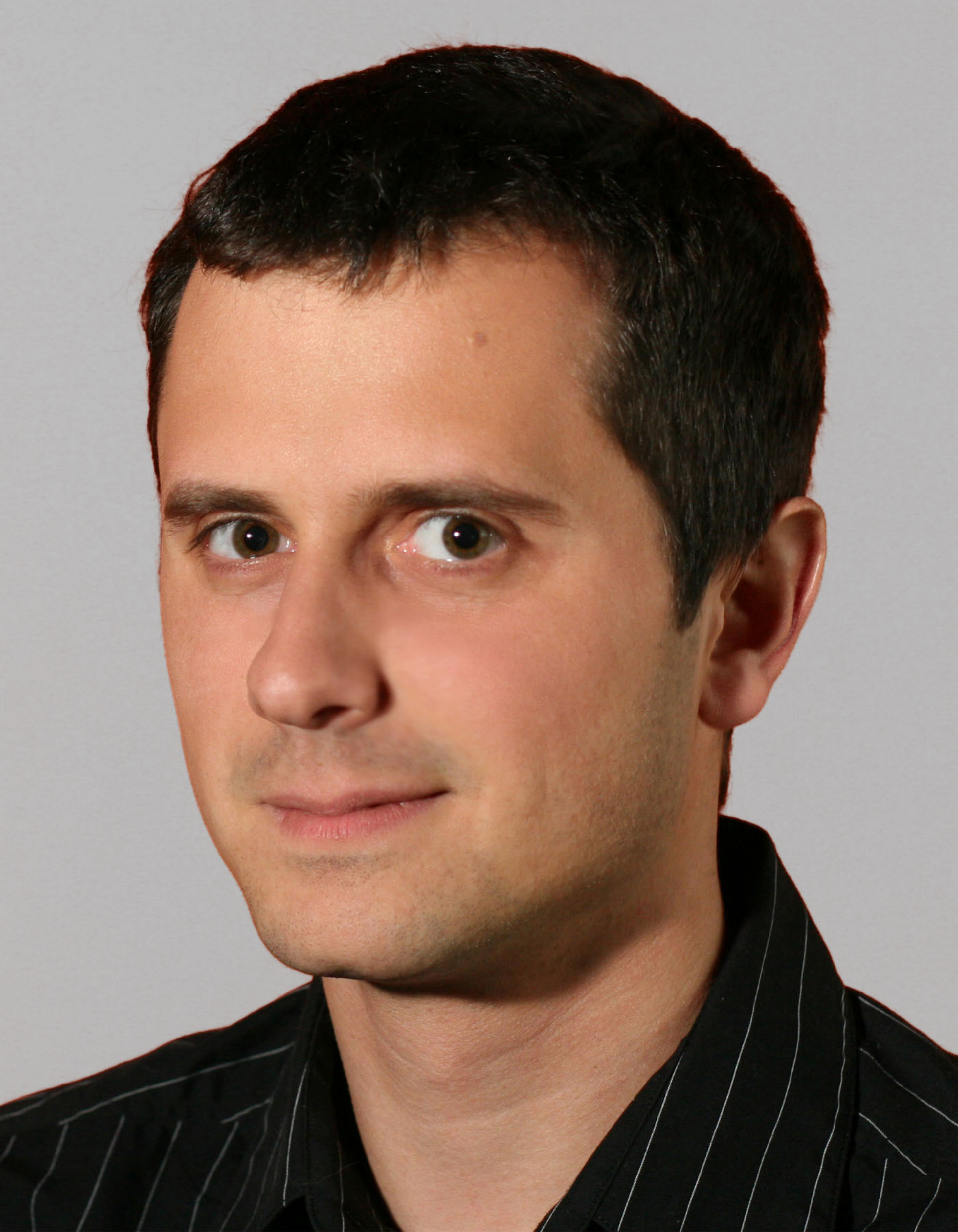}}]	{Sylvain Calinon} is a Senior Researcher at the Idiap Research Institute. He is also a lecturer at the Ecole Polytechnique F\'ed\'erale de Lausanne (EPFL), and an external collaborator at the Department of Advanced Robotics (ADVR), Italian Institute of Technology (IIT). From 2009 to 2014, he was a Team Leader at ADVR, IIT. From 2007 to 2009, he was a Postdoc at the Learning Algorithms and Systems Laboratory, EPFL, where he obtained his PhD in 2007. He is the author of 100+ publications in robot learning and human-robot interaction, with recognition including Best Paper Awards in the journal of Intelligent Service Robotics (2017) and at IEEE Ro-Man'2007, and Best Paper Award Finalist at ICRA'2016, ICIRA'2015, IROS'2013 and Humanoids'2009. He currently serves as Associate Editor in IEEE Transactions on Robotics (T-RO) and IEEE Robotics and Automation Letters (RA-L). Webpage: \texttt{\footnotesize http://calinon.ch}
	
\end{IEEEbiography} 

\begin{IEEEbiography}[{\includegraphics[width=1in,height=1.25in,clip,keepaspectratio]{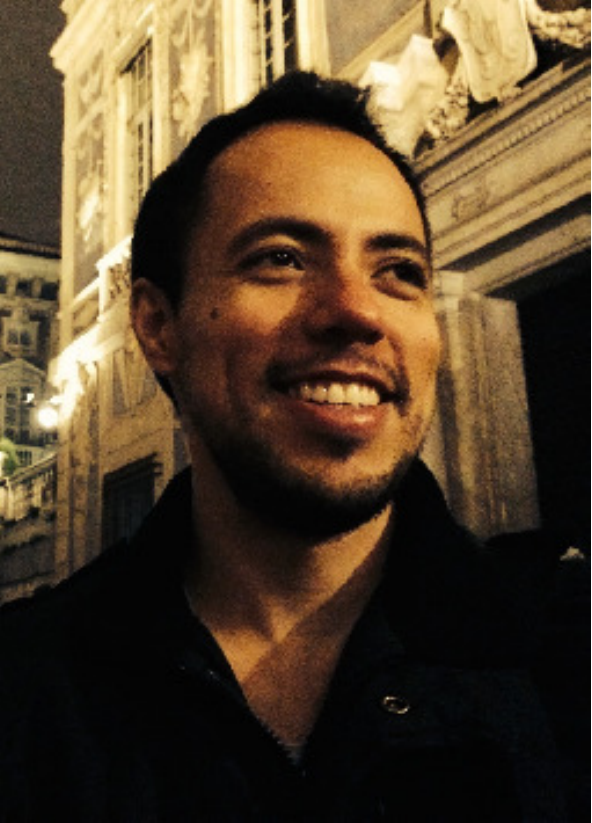}}]{Leonel Rozo}
	is a Team Leader at the Department of Advanced Robotics (ADVR), Istituto Italiano di Tecnologia since 2016. He was a postdoctoral researcher at the same institution from 2013 to 2016. He received his B.Sc on Mechatronics Engineering from the ''Nueva Granada" Military University (Colombia, 2005), his M.Sc in Automatic Control and Robotics (2007), and Ph.D in Robotics (2013) from the Polytechnical University of Catalonia (Barcelona, Spain). From 2007 to 2012 he carried out his research on force-based manipulation tasks learning at the Institut de Rob\`{o}tica i Inform\`{a}tica Industrial (CSIC-UPC). His research interests cover robot programming by demonstration, physical human-robot interaction, machine learning and optimal control for robotics. Personal webpage: \texttt{\footnotesize http://leonelrozo.weebly.com}
\end{IEEEbiography} 

\begin{IEEEbiography}[{\includegraphics[width=1in,height=1.25in,clip,keepaspectratio]{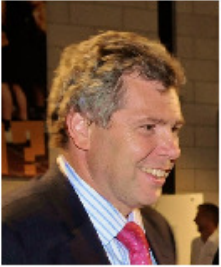}}]{Darwin G. Caldwell}\hspace{-0.1cm}, FREng is Deputy Director of the Italian Institute of Technology (IIT), and Director of the Dept. of Advanced Robotics at IIT. He is or has been an Honorary Professor at the Universities of Manchester, Sheffield, Bangor, Kings College London and Tianjin University, China. His research interests include; humanoid and quadrupedal robotics (iCub, cCub, COMAN, WalkMan, HyQ, HyQ2Max, HalfMan, COMAN+), innovative actuators, haptics and force augmentation exoskeletons, medical, rehabilitation and assistive robotic technologies, dexterous manipulators.  He is the author or co-author of over 500 academic papers, 20+ patents, and has received awards and nominations from many international journals and conferences. Caldwell has been chair of the IEEE Robotics and Automation Chapter (UKRI), a past co-chair of the IEE (IET) Robotics and Mechatronics PN, Secretary of the IEEE/ASME Trans. on Mechatronics, Editor for Frontiers in Robotics and AI, on the editorial board of the International Journal of Social Robotics and Industrial Robot and on the Advisory Board of Science Robotics. In 2015 he was elected a Fellow of the Royal Academy of Engineering.
%is a founding Director at the Italian Institute of Technology in Genoa, Italy, and a Honorary Professor at the Universities of Sheffield, Manchester, Bangor, Kings College, London and Tianjin University China. His research interests include innovative actuators, humanoid and quadrupedal robotics and locomotion (iCub, cCub, HyQ and COMAN), haptic feedback, force augmentation exoskeletons, dexterous manipulators, biomimetic systems, rehabilitation and surgical robotics, telepresence and teleoperation procedures. He is the author or co-author of over 450 academic papers, and 17 patents and has received awards and nominations from several international journals and conference including; IMechE Best Paper Award 2009, Ind. Robot Journal 2010, ICRA (2007), IROS (2007, 2012, 2013), ICAR (2003), Humanoids (2008, 2012), CASE (2008), ICMA (2011), Robio (2013) IFAC IAV, MMVR (2011), ACHI (2010), WorldHaptics (2007) and Virtual Concepts (2006). He is Editor for Frontiers in Robotics and AI, secretary of the IEEE/ASME Trans. on Mechatronics and on the editorial board of the International Journal of Social Robotics and Industrial Robot.
\end{IEEEbiography} 

\end{document}